%% file: main.tex
\pdfoutput=1

\documentclass[11pt]{article}

\usepackage[preprint]{acl}

\usepackage{times}
\usepackage{latexsym}

\usepackage[T1]{fontenc}

\usepackage[utf8]{inputenc}

\usepackage[final,nopatch=footnote]{microtype}

\usepackage{inconsolata}

\usepackage{graphicx}

\usepackage {mathtools,amssymb,amsthm} 
\usepackage{csquotes}
\usepackage{cleveref}
\usepackage{enumerate}
\usepackage{graphicx}
\usepackage{multirow}
\usepackage{tabularx}
\usepackage{array}
\usepackage{enumitem}
\usepackage[ruled,vlined]{algorithm2e}

\usepackage{booktabs}
\usepackage{tcolorbox}
\tcbuselibrary{most}
\usepackage{booktabs}
\usepackage{listings}
\usepackage[edges]{forest}
\usetikzlibrary{shadows}
\usepackage{caption}
\usepackage{lipsum} 
\usepackage[export]{adjustbox} 
\usepackage{setspace}        
\tcbuselibrary{skins,breakable} 

\newcommand{\para}[1]{\vspace{0.6mm}\noindent\textbf{#1}}

\definecolor{hbblue}{HTML}{000000}
\newcommand{\hb}[1]{{\color{black} #1}}

\title{CLLMate: A Multimodal Benchmark for Weather and Climate Events Forecasting}

\author{Haobo Li \and Zhaowei Wang \and Jiachen Wang \and YueYa Wang \\ {\bf Alexis Kai Hon Lau} \and {\bf Huamin Qu}\\
        Hong Kong University of Science and Technology \\ Hong Kong}

\begin{document}
\maketitle
\begin{abstract}
Forecasting weather and climate events is crucial for making appropriate measures to mitigate environmental hazards and minimize losses.
\hb{However, existing environmental forecasting research focuses narrowly on predicting numerical meteorological variables (e.g., temperature), neglecting the translation of these variables into actionable textual narratives of events and their consequences.}
To bridge this gap, we proposed \textbf{W}eather and \textbf{C}limate \textbf{E}vent \textbf{F}orecasting (WCEF), a new task that leverages numerical meteorological raster data and textual event data to predict weather and climate events. 
This task is challenging to accomplish due to difficulties in aligning multimodal data and the lack of supervised datasets.
\hb{To address these challenges, we present CLLMate, the first multimodal dataset for WCEF, using 26,156 environmental news articles aligned with ERA5 reanalysis data.
We systematically benchmark 23 existing MLLMs on CLLMate, including closed-source, open-source, and our fine-tuned models.
Our experiments reveal the advantages and limitations of existing MLLMs and the value of CLLMate for the training and benchmarking of the WCEF task.}

\end{abstract}

\input{1-intro.tex}
\input{2-rw.tex}
\input{3-pre.tex}
\input{4-dataset.tex}
\input{5-benchmark.tex}
\input{6-analysis.tex}
\input{7-conclusion.tex}
\input{8-limitation.tex}

\newpage
\bibliography{custom}

\appendix
\newpage
\onecolumn
\section{Dataset Release and License}
\subsection{CLLMate License and Availability.}
\label{data_format}
The CLLMate dataset will be publicly available after the review phase and use the CC BY 4.0 license. 
The data format is:
\colorlet{punct}{red!60!black}
\definecolor{background}{HTML}{EEEEEE}
\definecolor{delim}{RGB}{20,105,176}
\colorlet{numb}{magenta!60!black}

\lstdefinelanguage{json}{
    basicstyle=\normalfont\ttfamily,
    numbers=left,
    numberstyle=\scriptsize,
    stepnumber=1,
    numbersep=8pt,
    showstringspaces=false,
    breaklines=true,
    frame=lines,
    backgroundcolor=\color{background},
    literate=
     *{0}{{{\color{numb}0}}}{1}
      {1}{{{\color{numb}1}}}{1}
      {2}{{{\color{numb}2}}}{1}
      {3}{{{\color{numb}3}}}{1}
      {4}{{{\color{numb}4}}}{1}
      {5}{{{\color{numb}5}}}{1}
      {6}{{{\color{numb}6}}}{1}
      {7}{{{\color{numb}7}}}{1}
      {8}{{{\color{numb}8}}}{1}
      {9}{{{\color{numb}9}}}{1}
      {:}{{{\color{punct}{:}}}}{1}
      {,}{{{\color{punct}{,}}}}{1}
      {\{}{{{\color{delim}{\{}}}}{1}
      {\}}{{{\color{delim}{\}}}}}{1}
      {[}{{{\color{delim}{[}}}}{1}
      {]}{{{\color{delim}{]}}}}{1},
}

\begin{lstlisting}[language=json,firstnumber=1]
{
    ...
    "6973": {
        "caused by": "None",
        "cause": "6970",
        "news_id": "5838681",
        "event": "high temperature",
        "category_name": "phenomena.atmospheric.high temperature",
        "category_index": "A.A.A",
        "time": "2022/07/25",
        "location": "Guangzhou City",
        "coordinate": "[20.69, 110.88, 25.69, 115.88]",
        "image_path": "./image/6973.png",
        "temperature": {
                "max": "302.75",
                "min": "302.00",
                "mean": "302.50"
        },
        "wind speed": {
                "max": "6.56",
                "min": "4.09",
                "mean": "5.51"
        },
        "precipitation": {
                "max": "0.00",
                "min": "0.00",
                "mean": "0.00"
        }
    },
    "6974": {
        "caused by": "6979",
        "cause": "None",
        "news_id": "5677320",
        "event": "increased wildfire severity",
        "category_name": "consequence.disaster.fire",
        "category_index": "B.A.D",
        "time": "2022/07/25",
        "location": "California State",
        "coordinate": "[32.5288, -124.4096, 42.0095, -114.1312]",
        "image_path": "./image/6974.png",
        "temperature": {
                "max": "312.50",
                "min": "295.00",
                "mean": "306.00"
        },
        "wind speed": {

                "max": "12.07",
                "min": "0.00",
                "mean": "4.01"
        },
        "precipitation": {
                "max": "0.00",
                "min": "0.00",
                "mean": "0.00"
        }
    },
    ...
}
\end{lstlisting}


\subsection{Category}
\label{category}

\begin{forest}
  for tree={
      draw, semithick, rounded corners, 
        top color = white!40,
     bottom color = white!40,
       text width = 28mm, text badly centered,
             edge = {draw, semithick},
           anchor = east,
             grow = east,
    forked edge,
            s sep = 1mm,
            l sep = 8mm,    
         fork sep = 4mm,     
    where level>=1{          
      text width=36mm,       
      if level=1{            
        l sep=8mm,          
        fork sep=4mm
      }{}
    }{}
  }
  [Weather and Climate Events
    [Consequence, text width=24mm, top color=red!40, bottom color=red!40  
      [Transportation, text width=28mm, top color=red!40, bottom color=red!40
        [Impacted Maritime, top color=red!40, bottom color=red!40]
        [Impacted Road Traffic, top color=red!40, bottom color=red!40]
        [Impacted Train, top color=red!40, bottom color=red!40]
        [Impacted Flight, top color=red!40, bottom color=red!40]
      ]
      [Society Impact, text width=28mm, top color=red!40, bottom color=red!40
        [Impacted activity, top color=red!40, bottom color=red!40]
        [Impacted resource, top color=red!40, bottom color=red!40]
      ]
      [Damage, text width=28mm, top color=red!40, bottom color=red!40
        [Equipment Damage, top color=red!40, bottom color=red!40]
        [Vehicle Damage, top color=red!40, bottom color=red!40]
        [Building Damage, top color=red!40, bottom color=red!40]
        [Infrastructure Damage, top color=red!40, bottom color=red!40]
      ]
      [Organisms Impact, text width=28mm, top color=red!40, bottom color=red!40
        [Impacted Plant, top color=red!40, bottom color=red!40]
        [Impacted Animal, top color=red!40, bottom color=red!40]
        [Impacted Human, top color=red!40, bottom color=red!40]
      ]
      [Disaster, text width=28mm, top color=red!40, bottom color=red!40
        [Pollution, top color=red!40, bottom color=red!40]
        [Fire, top color=red!40, bottom color=red!40]
        [Earthquake \& Landslide, top color=red!40, bottom color=red!40]
        [Drought, top color=red!40, bottom color=red!40]
        [Flood, top color=red!40, bottom color=red!40]
      ]
    ]
    [Meteorological Phenomena, text width=24mm, top color=blue!30, bottom color=blue!30  
      [Wind, text width=28mm, top color=blue!30, bottom color=blue!30
        [Tornado, top color=blue!30, bottom color=blue!30]
        [Typhoon, top color=blue!30, bottom color=blue!30]
        [General Wind, top color=blue!30, bottom color=blue!30]
      ]
      [Precipitation, text width=28mm, top color=blue!30, bottom color=blue!30
        [Solid Precipitation, top color=blue!30, bottom color=blue!30]
        [Liquid Precipitation, top color=blue!30, bottom color=blue!30]
      ]
      [Atmospheric, text width=28mm, top color=blue!30, bottom color=blue!30
        [Low Visibility, top color=blue!30, bottom color=blue!30]
        [Low Temperature, top color=blue!30, bottom color=blue!30]
        [High Temperature, top color=blue!30, bottom color=blue!30]
      ]
    ]
  ]
\end{forest}

\newpage

\subsection{Category Distribution of Train and Test Sets.}
\label{distrubution}
\begin{figure}[ht!]
  \centering
  \includegraphics[width=\linewidth]{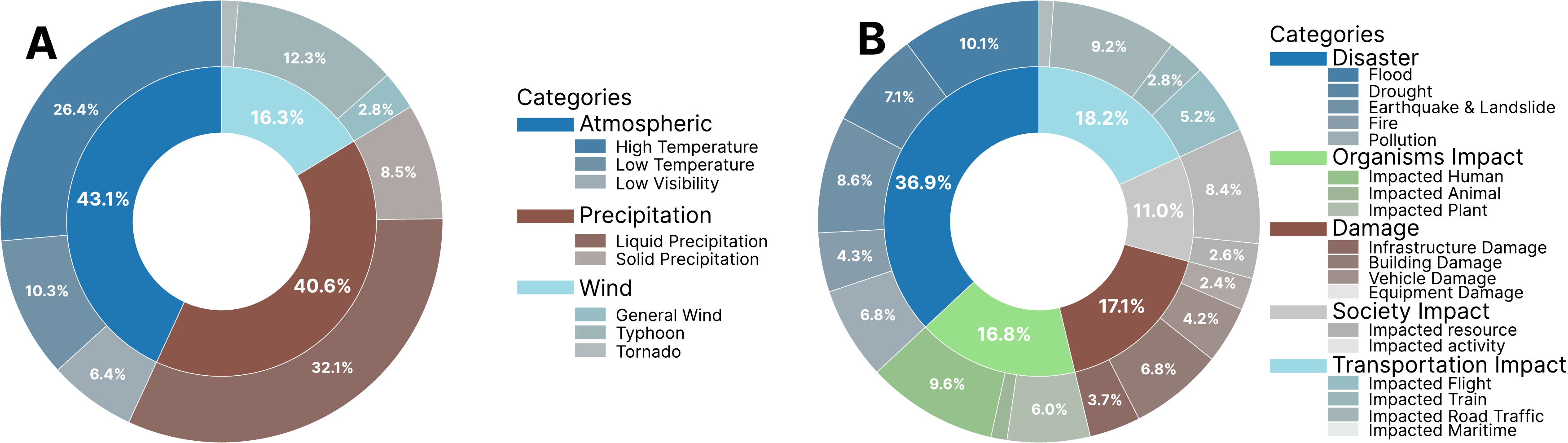}
  \caption{Distribution of categories of the training set. A: meteorological phenomena categories (3487/7747 events). B: consequences categories (3485/7747 events).}
  \label{fig:train}
\end{figure}

\begin{figure}[ht!]
  \centering
  \includegraphics[width=\linewidth]{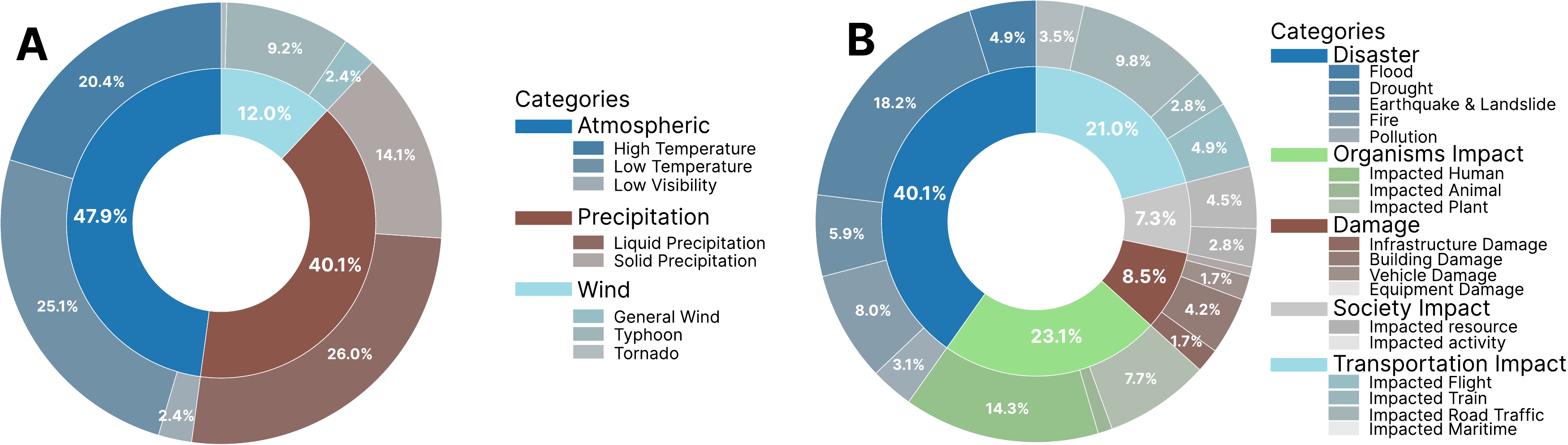}
  \caption{Distribution of categories of the testing set. A: meteorological phenomena categories (492/7747 events). B: consequences categories (283/7747 events).}
  \label{fig:test}
\end{figure}

The dataset is partitioned chronologically into training (90\%) and test (10\%) sets. We present the category distribution of training (\autoref{fig:train}) and testing (\autoref{fig:test}) sets. 

\subsection{Generalizability of CLLMate}
While the CLLMate benchmark is structured for hierarchical categorization via multiple-choice evaluation, its design inherently supports other discriminative or generative tasks, such as open-ended generation tasks. The inclusion of raw textual event descriptions (\autoref{data_format}) alongside aligned meteorological data enables applications beyond constrained categorization. For example:

\begin{itemize}[noitemsep,topsep=0pt,leftmargin=*]
    \item Free-form event narrative generation: Training or evaluating MLLMs to produce human-readable weather/climate event summaries from numerical inputs.

    \item Causal reasoning exploration: Probing model capabilities to infer and articulate chains of causality between meteorological drivers and societal impacts.

    \item Automated report drafting: Generating localized risk assessments or mitigation advisories grounded in spatio-temporal climate patterns.

    \item Open for integrating additional meteorological data (e.g., more meteorological variables) and more modalities (e.g., satellite images). For each event, we provide the location and date, making it easy to align future data.
\end{itemize}

The generalizability of CLLMate ensures compatibility with both discriminative and generative evaluation paradigms, broadening its utility for research in meteorology-informed language modeling and operational forecasting systems.

\subsection{ERA5 Reanalysis Dataset.}
The link to the ERA5 dataset is \url{https://cds.climate.copernicus.eu/datasets/reanalysis-era5-single-levels?tab=overview}.
The ERA5 dataset uses a license of Copernicus products. This license permits access to Copernicus Products for any lawful purpose. Authorized uses include, but are not limited to, reproduction, distribution, public dissemination, adaptation, modification, and integration with other data and information.

\noindent The meteorological variables of ERA5 used in CLLMate are:
\begin{tcolorbox}[title = {ERA5 Varaibles}]
\begin{itemize}[noitemsep,topsep=0pt,leftmargin=*]
    \item \textbf{2m temperature}: This parameter is the temperature of air at 2m above the surface of land, sea or inland waters. 2m temperature is calculated by interpolating between the lowest model level and the Earth's surface, taking account of the atmospheric conditions. This parameter has units of kelvin (K). Temperature measured in kelvin can be converted to degrees Celsius (°C) by subtracting 273.15.
    \item \textbf{10m u-component of wind}: This parameter is the eastward component of the 10m wind. It is the horizontal speed of air moving towards the east, at a height of ten metres above the surface of the Earth, in metres per second. This parameter can be combined with the V component of 10m wind to give the speed and direction of the horizontal 10m wind. This parameter has units of metre per second (m/s).
    \item \textbf{10m v-component of wind}: This parameter is the northward component of the 10m wind. It is the horizontal speed of air moving towards the north, at a height of ten metres above the surface of the Earth, in metres per second. This parameter can be combined with the U component of 10m wind to give the speed and direction of the horizontal 10m wind. his parameter has units of metre per second (m/s).
    \item \textbf{Total precipitation}: This parameter is the accumulated liquid and frozen water, comprising rain and snow, that falls to the Earth's surface. It is the sum of large-scale precipitation and convective precipitation. Large-scale precipitation is generated by the cloud scheme in the ECMWF Integrated Forecasting System (IFS). This parameter has units of meter (m).
    
\end{itemize}
\end{tcolorbox}

\subsection{The environmental news dataset.}
The environmental news dataset used to extract meteorological events was acquired through a procurement process from Wisers~\cite{wisers2024}.

\section{Prompt and Annotation}
\subsection{Prompt for Extracting Events}
\label{extract_prompt}
\tcbset{width=\textwidth}
\begin{tcolorbox}[title = {System Prompt}]
You are an AI assistant with expertise in extracting environmental events. Your task is to analyze the given news article and identify environmental events (and their consequences) and their corresponding locations and time that have actually occurred and are explicitly mentioned in the text. Only extract triples if they pertain to environmental events and actual events. Do not make inferences.
\end{tcolorbox}

\begin{tcolorbox}[title = {Output Definition}]
In the extracted triples, the first element should represent the subject, the second element should be the constrained verb 'cause,' and the third element should represent the object. Both the subject and object should relate to environmental events or risks. If the geographic location is not provided, indicate 'no'. If the date is provided, output YYYY/MM/DD. If the date is not provided, indicate 'no'.
\end{tcolorbox}

\begin{tcolorbox}[title = {Zero-Shot CoT}]
Initially, evaluate if the news article discusses environmental events. Next, determine if the associated geographic location and time are specified. If both conditions are met, output the list [subject, cause, object, location, time]. The output should be in English.
\end{tcolorbox}

\begin{tcolorbox}[title = {Few-shot}]
If the news is <news article containing environmental events>, the output should be <correct events with location and time>.
If the news is <news article containing no environmental events>, the output should be ``No''.
Now, the news is: <news article to be analyzed>
\end{tcolorbox}

\subsection{Annotation Details}
\label{annotation}
We presented the original news articles along with the extracted events to three domain experts for annotation. The experts are postdoc and PhD candidates in the domain.
Each expert assessed the extracted events based on the corresponding news article, assigning a label of 1 (accurate) or 0 (inaccurate). The final annotation for each event was determined by majority vote. Out of 6,352 news articles, the extracted events from 2,575 articles (40.54\%) were deemed accurate. The inter-annotator agreement (IAA) score, measured by the pairwise agreement proportion, was 81.83\%, and Fleiss’s $\kappa$~\cite{fleiss1971measuring} was calculated to be 0.63.

\subsection{Prompt for Phenomena Forecasting}
\label{phenomena_prompt}

\tcbset{width=\textwidth, colframe=blue!30}
\begin{tcolorbox}[title = {System Prompt}]
You are an AI assistant with expertise in weather analysis. Your task is to interpret meteorological data and figures, analyze weather variables, and provide accurate insights into weather conditions.
\end{tcolorbox}

\begin{tcolorbox}[title = {Location Information}]
The provided figure corresponds to <city\_name> city, covering the geographic area defined by:Latitude: [<lat\_min>, <lat\_max>], Longitude: [<lon\_min>, <lon\_max>].
\end{tcolorbox}

\begin{tcolorbox}[title = {Meteorological Parameters}]
The daily mean meteorological parameters I provided are as follows:

The 2m temperature: the temperature of air at 2m above the surface of land, sea or inland waters. This parameter has units of kelvin (K).
Average daily mean temperature of that area: <t\_mean> K. 
Max daily mean temperature of that area: <t\_max> K. 
Min daily mean temperature of that area: <t\_min> K. 

The 10m wind: the speed of air moving towards the north, at a height of ten metres above the surface of the Earth, in metres per second.
Average daily mean wind speed of that area: <w\_mean> m/s.
Max daily mean wind speed of that area: <w\_max> m/s.
Min daily mean wind speed of that area: <w\_min> m/s.

Precipitation: the accumulated liquid and frozen water, comprising rain and snow, that falls to the Earth's surface.  The units of this parameter are depth in metres of water equivalent.
Average daily mean precipitation of that area: <p\_mean> m.
Max daily mean precipitation of that area: <p\_max> m.
Min daily mean precipitation of that area: <p\_min> m.
\end{tcolorbox}

\begin{tcolorbox}[title = {Meteorological Image}]
The figure provided is an RGB image derived from spatiotemporal meteorological data for the region. Each channel represents:
channel 1: Temperature.
channel 2: Wind Speed.
channel 3: Precipitation.
\end{tcolorbox}

\begin{tcolorbox}[title = {Options (The order is random for different instances.)}]
The possible answers are structured into main categories with sub-options:

\begin{itemize}[noitemsep,topsep=0pt,leftmargin=*]
    \item A: Atmospheric
    \begin{itemize}[noitemsep,topsep=0pt,leftmargin=*]
    \item A.1: High Temperature
    \item A.2: Low Temperature
    \item A.3: Low Visibility
    \end{itemize}
    \item B: Wind
    \begin{itemize}[noitemsep,topsep=0pt,leftmargin=*]
    \item B.1: General Wind
    \item B.2: Tornado
    \item B.3: Typhoon
    \end{itemize}
    \item C: Precipitation
    \begin{itemize}[noitemsep,topsep=0pt,leftmargin=*]
    \item C.1: Solid Precipitation
    \item C.2: Liquid Precipitation
    \end{itemize}
\end{itemize}
\end{tcolorbox}

\begin{tcolorbox}[title = {Zero-Shot CoT}]
To determine the most likely meteorological scenario, follow these steps:
1. Analyze Statistical Data: Evaluate the statistical information (mean, max, and min values) for temperature, wind speed, and precipitation. Identify any conditions that suggest extreme weather or notable patterns.
2. Analyze Spatial Patterns in the Image: Examine the RGB image for regional variations in temperature, wind, and precipitation. Consider the spatial distribution of these variables to identify meteorological phenomena.
3. Synthesize Findings: Combine insights from the statistical data and image analysis.
Choose the only most appropriate sub-option from the given categories based on your analysis. You must output the serial number of the option.
\end{tcolorbox}

\subsection{Prompt for Consequences Forecasting}
\label{consequence_prompt}
\parbox{\textwidth}{For the prompt for the consequences forecasting, we will change the ``System Prompt'' and ``Options'' sections by using the corresponding categories.}

\tcbset{width=\textwidth, colframe=red!40}

\begin{tcolorbox}[title = {System Prompt}]
You are an AI assistant with expertise in weather analysis. Your task is to interpret meteorological data and figures, analyze weather variables, and provide accurate insights into meteorological consequences.
\end{tcolorbox}

\begin{tcolorbox}[title = {Options (The order is random for different instances.)}]
The possible answers are structured into main categories with sub-options:

\begin{itemize}[noitemsep,topsep=0pt,leftmargin=*]
    \item A: Disaster
    \begin{itemize}[noitemsep,topsep=0pt,leftmargin=*]
    \item A.1: Flood
    \item A.2: Drought
    \item A.3: Earthquake \& Landslide
    \item A.4: Fire
    \item A.5: Pollution, including Air Pollution, Water Pollution, Solid Pollution
    \end{itemize}
    \item B: Organisms Impact
    \begin{itemize}[noitemsep,topsep=0pt,leftmargin=*]
    \item B.1: Impacted Human, including Trap, Injury, Death
    \item B.2: Impacted Animal
    \item B.3: Impacted Plant
    \end{itemize}
    \item C: Damage
    \begin{itemize}[noitemsep,topsep=0pt,leftmargin=*]
    \item C.1: Infrastructure Damage
    \item C.2: Building Damage
    \item C.3: Vehicle Damage
    \item C.4: Equipment Damage
    \end{itemize}
    \item Society Impact
    \begin{itemize}[noitemsep,topsep=0pt,leftmargin=*]
    \item D.1: Impacted resource, including Food Shortage, Power Shortage
    \item D.2: Impacted activity, including Work Activity, School Activity
    \end{itemize}
    \item Transportation Impact
    \begin{itemize}[noitemsep,topsep=0pt,leftmargin=*]
    \item E.1: Impacted Flight
    \item E.2: Impacted Train
    \item E.3: Impacted Road Traffic
    \item E.4: Impacted Maritime
    \end{itemize}
\end{itemize}
\end{tcolorbox}

\section{Visual Instruction Tuning Models}
\label{training}
We fine-tuned the LLaVA models on the training set of CLLMate. The fine-tuned models include LLaVA-1.5-7B/13B and LLaVA-1.6-vicuna-7B/13B.
\subsection{Vision Encoder and Adapter}
We maintain consistency with LLaVA on the vision encoder and adapter.
For the vision encoder, we utilized CLIP ViT-L/14~\cite{radford2021learning} to extract the spatial patterns of meteorological images. For the vision-language adapter, we used the two-later MLP vision-language adapter provided by LLaVA.
The vision encoder and the adapter are frozen during the fine-tuning. 
\subsection{Training Configuration}
We fine-tuned LLaVA-7B and 13B models for one epoch using DeepSpeed ZeRO-3 on 8×A800 GPUs, with a learning rate of 2e-5 and batch size of 16. Training required 1 hour (7B) and 2 hours (13B) for one epoch.

\section{More Evaluation}
\label{analysis}
\subsection{Case Study}
\label{case}
To illustrate the model capabilities on the WCEF task, we analyze the representative examples (event 6973 and event 6974 in \autoref{data_format}) with the original news article and MLLMs' output.

\subsubsection{Case 1: Phenomena Forecasting (Event 6973)}
We present some typical MLLMs's output in \autoref{data_format}, with correct analyses highlighted in blue and incorrect ones in red, as assessed by a domain expert. 

Current MLLMs all provide structured, human-aligned rationales and identify meteorological thresholds (e.g., 29.35°C).
All models correctly excluded precipitation (C.1/C.2) and wind extremes (B.2/B.3); however, only Claude-3.5-Sonnet explicitly justified these exclusions using statistical ranges. The fine-tuned model, Fine-Tuned-LLaVA-1.6-vicuna-7B, provided direct answers without offering reasoning.

Regarding multimodal data, these models demonstrated accurate analysis of meteorological images. For instance, QWen2.5-VL-72B uniquely interpreted the meaning of colors, while LLaVA-34B concentrated on analyzing spatial uniformity.

\DeclareDocumentCommand{\degC}{}{%
   {\ifmmode\text{\textdegree C}%
    \else\textdegree C\fi}\xspace%
}

\newtcbtheorem{example}{Case}{
  breakable,
  colback=gray!5,
  colframe=gray!35!black,
  fonttitle=\bfseries}{x}

\begin{example}{Event 6973 (shown in subsection A.1)}{}
\label{6973}
\textbf{The news article: }\\
Guangzhou issued its first high temperature red warning this year. Guangzhou will continue to be sunny and hot in the next three days.\\
Southern Metropolis Daily\\
Original Report | July 25, 2022, 16:24\\

Guangzhou has issued its first red high-temperature warning of the year. As of 4 p.m., seven districts in Guangzhou, excluding Conghua, Zengcheng, Nansha, and Panyu, have issued red warnings for high temperatures. The highest temperatures in these districts are expected to reach around 39°C today and tomorrow. According to the provincial government's defense guidelines, outdoor activities should be minimized during the daytime, and outdoor work in open areas should be suspended, except for special industries.

Under the influence of a strong subtropical high-pressure system, Guangzhou is expected to experience widespread sunny and hot weather with high temperatures over the next three days. 
\noindent\rule{\textwidth}{0.5pt}
\medskip 
\begin{minipage}[t]{0.68\linewidth}
\vspace*{0pt} 
Ground Truth: A.1: High Temperature (option in the question)\\
Event: high temperature\\
Category\_name: phenomena.atmospheric.high temperature\\
Category\_index: A.A.A\\
Time: 2022/07/25\\
Location: Guangzhou City\\
\end{minipage}\hfill
\begin{minipage}[t]{0.28\linewidth}
\vspace*{0pt}
\centering
\includegraphics[width=\linewidth,height=0.1\textheight,keepaspectratio]{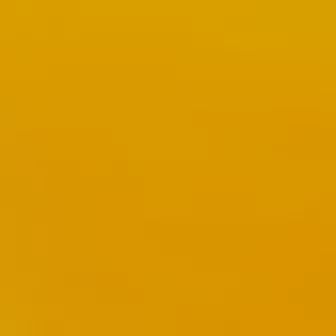} 
\label{fig:event6973} 
\end{minipage}

\tcblower
\textbf{GPT-4o's output:}
\begin{itemize}[noitemsep,topsep=0pt,leftmargin=*]
    \item 1. Analyze Statistical Data:
    \begin{itemize}
        \item Temperature: Average daily mean temperature is 302.50 K, with a max of 302.75 K and a min of 302.00 K. These values suggest relatively high temperatures.
        \item Wind Speed: Average daily mean wind speed is 5.51 m/s, with a max of 6.56 m/s and a min of 4.09 m/s. These figures suggest moderate wind speeds.
        \item Precipitation: All values show 0.00 m, indicating no precipitation.
    \end{itemize}
    
    \item 2. Analyze Spatial Patterns in the Image: \\
    The image predominantly shows \textcolor{blue}{a strong red hue, indicating high temperatures. There is also a noticeable green component, indicating the presence of notable wind speeds.}
    \item 3. Synthesize Findings:\\
    The data indicates \textcolor{blue}{high temperatures without any precipitation and moderate winds}.
\end{itemize}
Conclusion: The most appropriate option based on the analysis is: \textcolor{blue}{A.1: High Temperature}.

\vspace{0.4em}
\noindent\rule{\textwidth}{0.5pt}
\vspace{0.4em}
\textbf{Claude-3.5-Sonnet's output:}\\
Let me analyze the meteorological data and image systematically:
\begin{itemize}[noitemsep,topsep=0pt,leftmargin=*]
    \item 1. Statistical Data Analysis:
    \begin{itemize}
        \item Temperature: ~302.5K (29.35\degC) with small variation (0.75K range)
        \item Wind speed: Average 5.51 m/s with moderate variation (2.47 m/s range)
        \item Precipitation: 0.00m (no precipitation)
    \end{itemize}
    \item 2. Image Analysis: The image shows a predominantly yellow/orange color. \textcolor{blue}{High intensity in the red channel (temperature). Moderate intensity in green channel (wind). No visible blue component (confirming zero precipitation).}
    \item 3. Synthesis: The high temperature (302.5K/29.35\degC) and \textcolor{blue}{uniform yellow/orange coloring} indicates hot conditions. Wind speeds are moderate but not extreme. No precipitation is present. The spatial pattern shows \textcolor{blue}{consistent high temperature across the region}. Based on these observations: The dominant feature is the high temperature. No extreme wind events (ruling out B options). No precipitation (ruling out C options). \textcolor{blue}{The uniform high temperature pattern best matches A.1.}
\end{itemize}
Therefore, I select option \textcolor{blue}{A.1 (High Temperature)} as the most appropriate classification for these conditions. Answer: A.1

\vspace{0.4em}
\noindent\rule{\textwidth}{0.5pt}
\vspace{0.4em}
\textbf{LLaVA-1.6-vicuna-34B's output:}\\
Based on the provided meteorological data and the RGB image, the most likely meteorological scenario for the region is: \textcolor{blue}{A: Atmospheric. A.1: High Temperature.} The average daily mean temperature of 302.50 K (29.50\degC) is relatively high, and the maximum daily mean temperature of 302.75 K (29.75\degC) is even higher. The image shows a \textcolor{blue}{uniform distribution of high temperatures across the region, with no significant variations in temperature.} The wind speed and precipitation are relatively low and do not show any extreme patterns. The high temperatures suggest a warm and sunny climate, which is consistent with the image's depiction of high temperatures.

\vspace{0.4em}
\noindent\rule{\textwidth}{0.5pt}
\vspace{0.4em}
\textbf{Qwen2.5-VL-72B's output:}\\
A.1. Analysis:\\
\begin{itemize}[noitemsep,topsep=0pt,leftmargin=*]
    \item Step 1: Analyze Statistical Data
    \begin{itemize}
        \item Temperature: Average daily mean temperature: 302.50 K (approximately 29.35\degC). Max daily mean temperature: 302.75 K (approximately 29.60\degC). Min daily mean temperature: 302.00 K (approximately 28.85\degC). These temperatures indicate relatively high and consistent warmth across the region, suggesting a hot day.
        \item Wind Speed: Average daily mean wind speed: 5.51 m/s. Max daily mean wind speed: 6.56 m/s. Min daily mean wind speed: 4.09 m/s. The wind speeds are moderate but not indicative of extreme weather like tornadoes or typhoons. 
        \item Precipitation: Average daily mean precipitation: 0.00 m. Max daily mean precipitation: 0.00 m. Min daily mean precipitation: 0.00 m. There is no recorded precipitation, indicating dry conditions.
    \end{itemize}
    \item Step 2: Analyze Spatial Patterns in the Image. \textcolor{blue}{The image is uniformly yellow, which suggests a consistent pattern across the region.} Given the color coding: Red (Temperature): \textcolor{blue}{The yellow hue indicates a high temperature since red is associated with temperature. Green (Wind Speed): The lack of green suggests moderate wind speeds, consistent with the statistical data. Blue (Precipitation): The absence of blue confirms the lack of precipitation.}
    \item Step 3: Synthesize Findings. Combining the statistical data and the spatial patterns observed in the image, \textcolor{blue}{it is clear that the region experienced high temperatures with no significant precipitation and moderate wind speeds.}
\end{itemize}
Therefore, the most appropriate sub-option is \textcolor{blue}{A.1: High Temperature.}

\vspace{0.4em}
\noindent\rule{\textwidth}{0.5pt}
\vspace{0.4em}
\textbf{Fine-Tuned-LLaVA-1.6-vicuna-7B's output:}\\
\textcolor{blue}{A.1: High Temperature}
\end{example}

\subsubsection{Case 2: Consequence Forecasting (Event 6974)}
We present some typical MLLMs's output in \autoref{data_format}, with correct analyses highlighted in blue and incorrect ones in red, as assessed by a domain expert. 

This case demonstrates critical disparities in multimodal causal reasoning for fire risk prediction. While GPT-4o and Claude-3.5-Sonnet identified fire-prone conditions (high temperatures, moderate winds (should be strong wind), and zero precipitation), both erroneously selected A.4: Drought rather than A.5: Fire. In contrast, Qwen2.5-VL-72B successfully synthesized all variables, especially interpreting wind spikes as fire accelerants, to correctly predict A.5: Fire. 

\begin{example}{Event 6974 (shown in subsection A.1)}{}
\label{6974}
\textbf{The news article: }\\
Wildfire spreads near Yosemite National Park in the United States\\

On July 25 local time, the spread of a wildfire near Yosemite National Park in California slowed down, the containment rate increased, and firefighting efforts showed signs of progress.

The California Department of Forestry and Fire Protection (Cal Fire) stated on the evening of the 25th that it was a "successful day" for firefighting efforts. From morning to evening, strong winds caused the burned area to expand from 16,791 acres (approximately 68 square kilometers) to 17,241 acres (approximately 70 square kilometers), while the containment rate rose from 10\% to 16\%. The agency reported that the fire had destroyed 55 structures. As firefighting efforts advanced, evacuation orders in some areas were downgraded to evacuation warnings, and certain roads were reopened for local residents.

On the 25th, a China News Service reporter observed at the fire site that firefighters had cleared a large number of firebreaks through the forested areas. In areas near major roads, the fire was no longer severe, but thick smoke continued to rise from the hillsides and mountain tops. Several aircraft took turns dropping water or fire retardants onto larger hotspots, while fire trucks were dispersed across various locations to respond to potential flare-ups. On the charred land, small groups of firefighters were extinguishing smaller fires.

According to Cal Fire, as of the evening of the 25th, nearly 3,000 firefighters, 24 helicopters, 302 fire engines, and 68 water trucks had been deployed to the fire site. On that day alone, helicopters dropped 300,000 gallons (approximately 1.136 million liters) of water onto the fire.

With climate change causing increasingly hotter and drier conditions in the western United States, wildfires in California have become more severe. This wildfire near Yosemite National Park started in the Midpines area of Mariposa County and spread rapidly due to strong winds, becoming California's largest wildfire so far this year. U.S. media have reported that this wildfire is likely the beginning of another severe wildfire season in California.
\noindent\rule{\textwidth}{0.5pt}
\medskip %
\begin{minipage}[t]{0.68\linewidth}
\vspace*{0pt} 
Ground Truth: A.5: Fire (option in the question)\\
Event: increased wildfire severity\\
Category\_name:consequence.disaster.fire \\
Category\_index: B.A.D\\
Time: 2022/07/25\\
Location: California State \\
\end{minipage}\hfill
\begin{minipage}[t]{0.28\linewidth}
\vspace*{0pt}
\centering
\includegraphics[width=\linewidth,height=0.1\textheight,keepaspectratio]{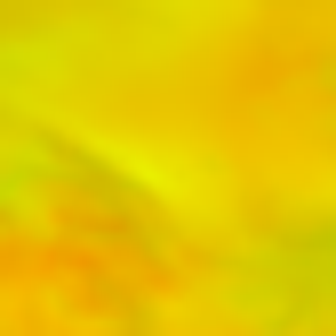} %
\label{fig:event6974} %
\end{minipage}
\tcblower
\textbf{GPT-4o's output:}
To determine the most likely meteorological scenario, let's analyze the provided data and image:
\begin{itemize}[noitemsep,topsep=0pt,leftmargin=*]
    \item 1. Analyze Statistical Data:
    \begin{itemize}
        \item Temperature: The average daily mean temperature is 306.00 K, with a max of 312.50 K. This indicates high temperatures, potentially indicative of heat waves.
        \item Wind Speed: The average daily mean wind speed is 4.01 m/s with peaks at 12.07 m/s. This suggests mild to moderate winds, with occasional stronger gusts.
        \item Precipitation: There is no precipitation recorded (0.00 m), indicating very dry conditions.
    \end{itemize}
    \item 2. Analyze Spatial Patterns in the Image:
    The image is predominantly \textcolor{blue}{yellow with reddish tones, indicating high temperatures (red channel is for temperature)}. This suggests a uniform region of warm temperatures. The lack of blue and green tones indicates negligible precipitation and moderate to \textcolor{red}{minimal wind activity.}
    \item 3. Synthesize Findings:
    Temperature and Precipitation: The very high temperature coupled with the lack of precipitation suggests a dry heat scenario. Impact Analysis: The conditions are suggestive of potential drought and \textcolor{blue}{fire risks due to high temperatures and no rainfall}. Considering the provided categories: Drought (A.4): \textcolor{red}{Prolonged high temperatures} and no precipitation are indicative of drought conditions, impacting water availability, agriculture, and potentially leading to wildfires.
\end{itemize}
Given this analysis, the most appropriate sub-option based on the scenario is: \textcolor{red}{A.4: Drought.}

\vspace{0.4em}
\noindent\rule{\textwidth}{0.5pt}
\vspace{0.4em}
\textbf{Claude-3.5-Sonnet's output:}\\
Let me analyze the meteorological data and image systematically:
\begin{itemize}[noitemsep,topsep=0pt,leftmargin=*]
    \item 1. Statistical Data Analysis:
    \begin{itemize}
        \item Mean: 306.00 K (32.85\degC). Max: 312.50 K (39.35\degC). Min: 295.00 K (21.85\degC). These are very high temperatures, indicating extreme heat conditions.
        \item Wind: Mean: 4.01 m/s. Max: 12.07 m/s. Min: 0.00 m/s.  Moderate to high wind speeds present.
        \item Precipitation: All values at 0.00 m.  Indicating completely dry conditions with no rainfall.
    \end{itemize}
    \item 2. Image Analysis: \textcolor{blue}{Dominant yellow-orange coloring indicates high temperatures. Some variations in green suggesting wind speed variations. No blue tinting, confirming zero precipitation.}
    \item 3. Synthesis: The combination of: \textcolor{blue}{Extremely high temperatures (approaching 40\degC). Complete absence of precipitation. Presence of wind. Geographic location (California).} These conditions strongly indicate a severe drought situation with \textcolor{blue}{high fire risk}. The complete lack of precipitation, combined with high temperatures and wind, creates perfect conditions for drought conditions.
\end{itemize}
Based on the given options and this analysis, the most appropriate classification would be: \textcolor{red}{A.4 (Drought)}. The conditions present clear indicators of drought conditions, which can lead to \textcolor{blue}{water shortages and increased fire risk}. This is supported by both the statistical data showing zero precipitation and very high temperatures, as well as the image showing warm conditions across the region.

\vspace{0.4em}
\noindent\rule{\textwidth}{0.5pt}
\vspace{0.4em}

\textbf{LLaVA-1.6-vicuna-34B's output:}\\
Based on the provided meteorological data and the RGB image, I will analyze the situation and provide the most appropriate sub-option from the given categories.
\begin{itemize}
    \item 1. Analyze Statistical Data: The average daily mean temperature of the area is 306.00 K, which is equivalent to 32.83 degrees Celsius. The max daily mean temperature is 312.50 K (34.38 degrees Celsius), and the min daily mean temperature is 295.00 K (27.27 degrees Celsius). The average daily mean wind speed is 4.01 m/s, with a max daily mean wind speed of 12.07 m/s and a min daily mean wind speed of 0.00 m/s. The average daily mean precipitation is 0.00 m, with a max daily mean precipitation of 0.00 m and a min daily mean precipitation of 0.00 m. From the statistical data, we can observe that the area experiences relatively high temperatures, with a range of 7.56 degrees Celsius between the max and min daily mean temperatures. The wind speeds are relatively low, with a range of 12.07 m/s between the max and min daily mean wind speeds. The precipitation levels are very low, with no precipitation recorded.
    \item 2. Analyze Spatial Patterns in the Image: The RGB image shows variations in temperature, wind speed, and precipitation across the region. The red channel (temperature) shows a gradient of temperature, with warmer areas in the south and cooler areas in the north. The green channel (wind speed) shows a gradient of wind speed, with higher wind speeds in the south and lower wind speeds in the north. The blue channel (precipitation) shows no precipitation across the region. From the image analysis, we can observe that the region experiences \textcolor{blue}{a gradient of temperature and wind speed, with warmer and windier conditions in the south and cooler and calmer conditions in the north. However, there is no precipitation recorded in the image.}
    \item 3. Synthesize Findings: Based on the statistical data and image analysis, the most appropriate sub-option from the given categories is: \textcolor{red}{D: Organisms Impact D.1: Impacted Plant.} The reason for this choice is that the region experiences high temperatures and low precipitation, which can lead to drought conditions. Drought can have a significant impact on plant life, as it can cause stress, wilting, and even death. The lack of precipitation in the image further supports this conclusion.
\end{itemize}
Therefore, the serial number of the most appropriate sub-option is: D.1.

\vspace{0.4em}
\noindent\rule{\textwidth}{0.5pt}
\vspace{0.4em}
\textbf{Qwen2.5-VL-72B's output:}\\
Based on the provided meteorological data and the analysis of the RGB image:\\
\begin{itemize}[noitemsep,topsep=0pt,leftmargin=*]
    \item Step 1: Analyze Statistical Data
    \begin{itemize}
        \item Temperature: Average daily mean temperature: 306.00 K (approximately 32.85\degC). Maximum daily mean temperature: 312.50 K (approximately 39.35\degC). Minimum daily mean temperature: 295.00 K (approximately 21.85\degC). These temperatures indicate relatively warm conditions, with some areas experiencing high temperatures.
        \item Wind Speed: Average daily mean wind speed: 4.01 m/s. Maximum daily mean wind speed: 12.07 m/s. Minimum daily mean wind speed: 0.00 m/s. The average wind speed is moderate, but there are areas with higher wind speeds, which could be significant.
        \item Precipitation: Average daily mean precipitation: 0.00 m. Maximum daily mean precipitation: 0.00 m. Minimum daily mean precipitation: 0.00 m. There is no recorded precipitation, indicating dry conditions.
    \end{itemize}
    \item Step 2: Analyze Spatial Patterns in the Image. \textcolor{blue}{Temperature (Red Channel): The red color dominates the image, suggesting widespread high temperatures across the region. Wind Speed (Green Channel): The green color is present but less dominant compared to red, indicating variable wind speeds with some areas experiencing higher wind speeds. Precipitation (Blue Channel): The absence of blue suggests no precipitation across the region.}
    \item Step 3: Synthesize Findings. Given the high temperatures, lack of precipitation, and variable wind speeds, \textcolor{blue}{the most likely meteorological scenario is one associated with dry and hot conditions. The combination of high temperatures and wind can lead to increased risk of wildfires.}
\end{itemize}
Conclusion: The most appropriate sub-option based on the analysis is: \textcolor{blue}{A.5: Fire.}

\vspace{0.4em}
\noindent\rule{\textwidth}{0.5pt}
\vspace{0.4em}
\textbf{Fine-Tuned-LLaVA-1.6-vicuna-7B's output:}\\
\textcolor{red}{A.2: Earthquake \& Landslide}
\end{example}

\subsection{Ablation Study}
\label{ablation}

To quantify the contribution of multimodal inputs, we conducted the ablation study with Gemini-2.0-flash (closed-source) and LLaVA-1.6-mistral-7B (open-source) for their good performance (e.g., LLaVA-1.6-mistral-7B has the best performance among 7/8B open-source MLLMs). 
We ablated three components: meteorological images, statistical summaries, and Zero-Shot CoT prompting. The results are displayed in \autoref{tab:ablation}.

\begin{table*}[ht!]
\centering
  \setlength{\aboverulesep}{1pt}
\setlength{\belowrulesep}{1pt}
\makeatletter
\def\hlinewd#1{%
  \noalign{\ifnum0=`}\fi\hrule \@height #1 \futurelet
   \reserved@a\@xhline}
\makeatother
\newcolumntype{C}[1]{>{\centering\arraybackslash}p{#1}}
\begin{tabular}{ p{6.2cm} | C{1.8cm}C{1.8cm} | C{2cm}C{2cm}}
\hlinewd{0.8pt}
\toprule
Models & \small{Primary Category of Phenomena} & \small{Subordinate Category of Phenomena} & \small{Primary Category of Consequence} & \small{Subordinate Category of Consequence} \\
\midrule
Gemini-2.0-flash~\cite{team2023gemini} & 37.80\% & 19.72\% & 30.39\% & 13.43\% \\ 
\hspace{1em} w/o image & 25.81\% & 14.63\% & 30.74\% & 14.84\%  \\ 
\hspace{1em} w/o statistics & 37.80\% & 17.48\% & 38.87\% & 13.43\%  \\ 
\hspace{1em} w/o Zero-Shot CoT & 33.94\% & 11.99\% & 17.67\% & 3.89\%  \\ 
\midrule
LLaVA-1.6-mistral-7B~\cite{liu2024llavanext} & 32.93\% & 17.28\% & 39.22\% & 12.01\%  \\ 
\hspace{1em} w/o image & 27.64\% & 13.41\% & 13.78\% & 4.95\%  \\ 
\hspace{1em} w/o statistics & 26.22\% & 15.45\% & 44.52\% & 7.77\%  \\ 
\hspace{1em} w/o Zero-Shot CoT & 33.33\% & 10.77\% & 20.49\% & 3.89\%  \\ 
\bottomrule
\end{tabular}
\caption{The results of the ablation study show the accuracy of benchmarked MLLMs across two granularity levels: primary category classification (coarse-grained) and subordinate category identification (fine-grained) for both meteorological phenomena and their cascading consequences. w/o image: remove the meteorological image from the prompt. w/o: remove the statistics of meteorological variables from the prompt. w/o: remove the Zero-Shot CoT from the prompt.}
\label{tab:ablation}
\end{table*}

\subsubsection{Closed-Source Models (Gemini-2.0-Flash)}
\noindent\textbf{Image Removal:} Eliminating meteorological raster data catastrophically degrades phenomena forecasting (37.80\% → 25.81\%, 19.72\% → 14.63\%),  but marginally improves consequence subordinate accuracy (30.39\% → 30.74\%, 13.43\% → 14.84\%). This suggests Gemini-2.0-flash relies heavily on spatial patterns for phenomena forecasting.

\noindent\textbf{Statistics Removal:} Removing statistics surprisingly does not significantly impact performance, improving overall accuracy from 30.39\% to 38.87\% of primary consequence forecasting. This indicates a redundancy between statistical summaries and raster data for Gemini-2.0-flash.

\noindent\textbf{Zero-Shot CoT Removal:} Disabling Zero-Shot CoT harms event forecasting, especially for consequence forecasting for Gemini-2.0-flash, underscoring its necessity for causal event differentiation.

\subsubsection{Open-Source Models (LLaVA-1.6-Mistral-7B)}
\noindent\textbf{Image Removal:} Degrades performance universally and largely, with catastrophic drops in consequence forecasting (39.22\% → 13.78\%). Unlike Gemini-2.0-flash, LLaVA-1.6-mistral-7B cannot compensate for missing spatial data via text-based statistics.

\noindent\textbf{Statistics Removal:} Improves primary consequence accuracy (39.22\% → 44.52\%) while harming phenomena prediction (32.93\% → 26.22\%), suggesting statistical summaries introduce noise for impact forecasting but aid event detection for LLaVA-1.6-mistral-7B.

\noindent\textbf{Zero-Shot CoT Removal:} Erases Zero-Shot CoT, reducing the performance of LLaVA-1.6-mistral-7B with the exception of primary phenomena forecasting.

\subsubsection{Key Insights}
\noindent\textbf{Modality Asymmetry:} MLLMs exhibit flexible modality compensation (e.g., statistics → images) in part of tasks but fail to compensate for all tasks. For example, LLaVA-1.6-mistral-7B has a performance drop for primary phenomena forecasting but not for consequence forecasting when ablating statistics.

\noindent\textbf{Reasoning Bottleneck:} Zero-Shot CoT prompting provides gains, especially in subordinate accuracy, proving essential for parsing causal meteorology-event relationships.

\subsection{Impact of Model Scale}

\begin{figure}[ht!]
  \centering
  \includegraphics[width=\linewidth]{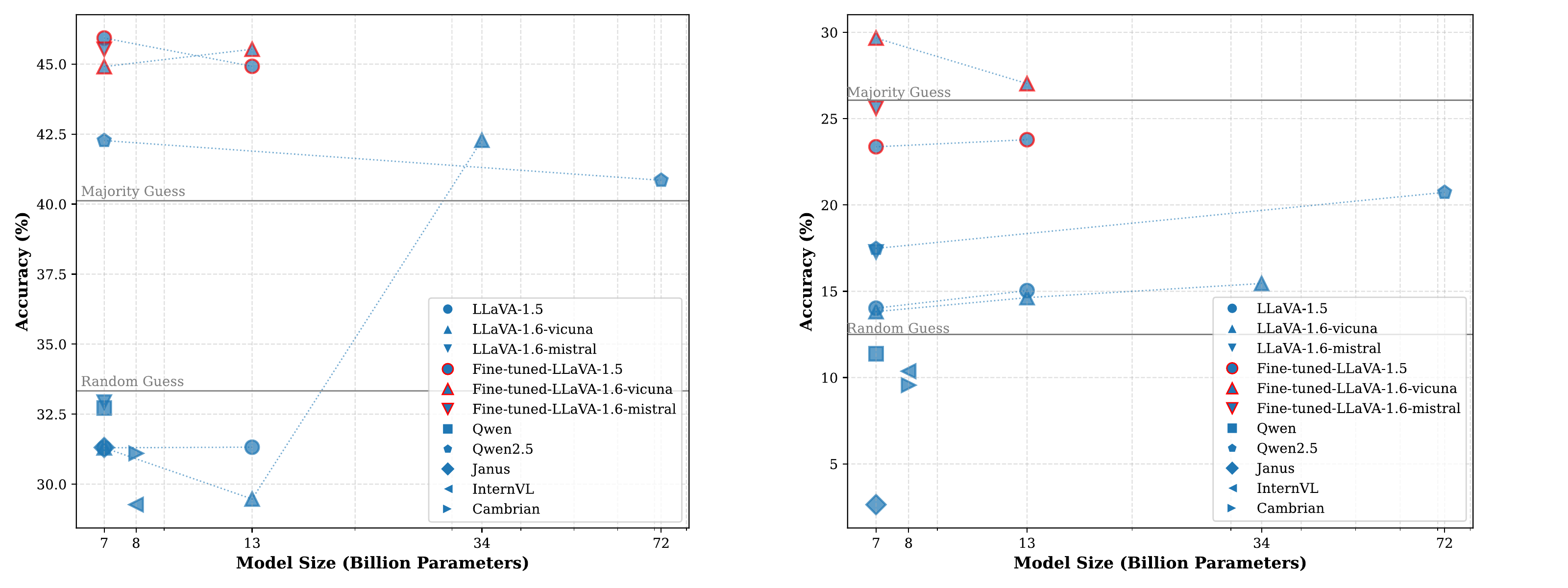}
  \caption{The line chart of the accuracy of open-sourced MLLMs. Left: Primary category of phenomena forecasting. Right: Subordinate category of phenomena forecasting. The x-axis is in the log scale.}
  \label{fig:A}
\end{figure}

\begin{figure}[ht!]
  \centering
  \includegraphics[width=\linewidth]{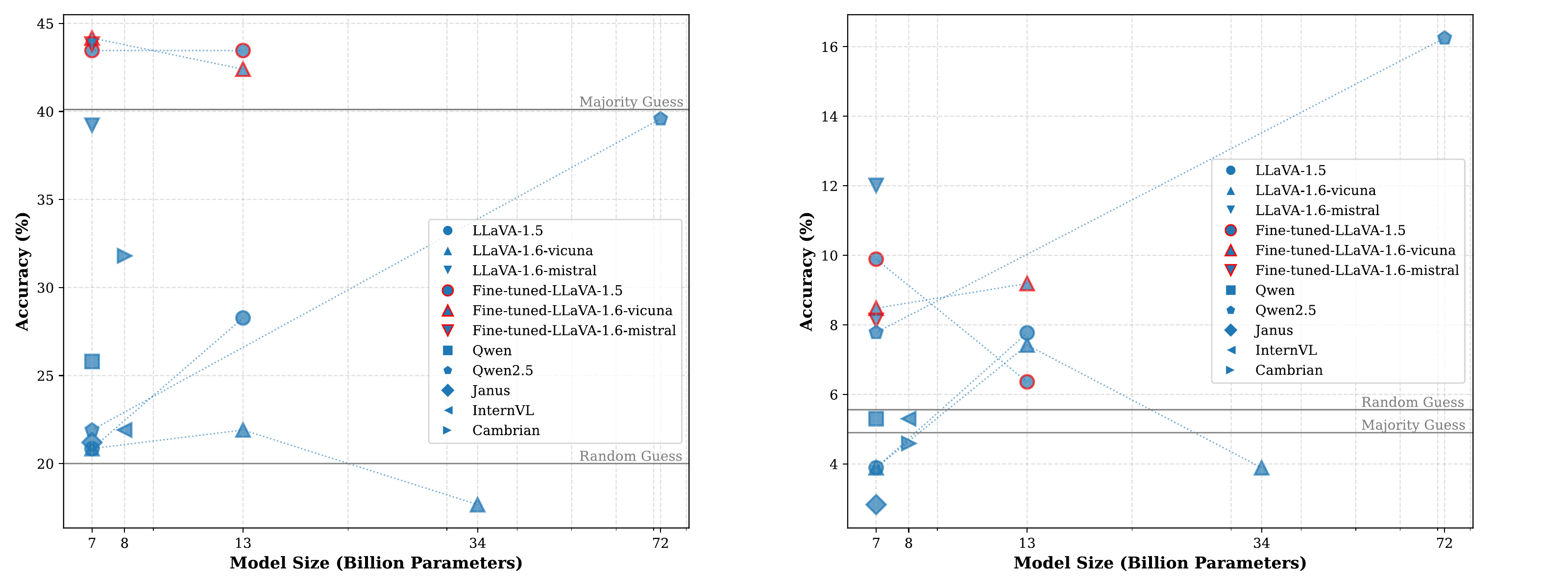}
  \caption{The line chart of the accuracy of open-sourced MLLMs. Left: Primary category of consequence forecasting. Right: Subordinate category of consequence forecasting. The x-axis is in the log scale.}
  \label{fig:B}
\end{figure}

We conducted the scale study on the open-source models, the results are shown in \autoref{fig:A} and \autoref{fig:B}.
\label{sec:scale }


\subsubsection{Scale}
Contrary to expectations, increasing the model scale does not guarantee improved performance:

LLaVA-1.6-34B achieves state-of-the-art primary phenomena accuracy (42.28\%) but catastrophically fails at primary consequence forecasting (17.67\% vs. 20.85\% for its 7B variant), suggesting larger models overfit to meteorological patterns while losing causal reasoning capabilities of consequence.

The Qwen2.5-VL series demonstrates scale-dependent tradeoffs:
Scaling from 7B to 72B improves subordinate consequence accuracy by 17.67\% in primary consequence forecasting (21.91\% → 39.58\%) and 8.47\% in subordinate consequence forecasting (7.78\% → 16.25\%) but reduces primary phenomena forecasting (42.27\% → 40.85\%).

\subsubsection{Architectural Refinements}
Architectural improvements (LLaVA-1.5 → 1.6, Qwen2 → Qwen2.5) yield significant gains.
The LLaVA-1.6-Mistral-7B variant outperforms LLaVA-1.5-7B in all tasks. The Qwen2.5-VL-7B outperforms the Qwen2-VL-7B except for the primary consequence forecasting.

\subsubsection{Fine-Tuning as a Scaling Alternative}
Task-specific adaptation eclipses scale advantages:
Smaller fine-tuned models consistently outperform larger untrained counterparts (13B/34B), with Fine-tuned-LLaVA-1.6-vicuna-7B surpassing the 34B untrained variant by 26.50\% in primary consequence forecasting (44.17\% vs. 17.67\%).
Fine-tuned-LLaVA-1.6-vicuna-7B also achieves 29.67\% subordinate phenomena accuracy vs. 15.45\% for LLaVA-1.6-vicuna-34B.
Fine-tuned 7B models outperform larger untrained counterparts, proving task alignment outweighs raw scale for meteorology-to-text translation.

\end{document}

%% file: 1-intro.tex
\section{Introduction}
Weather and climate events, namely discrete episodes of extreme weather or abnormal climate conditions~\cite{hurrell2013climate}, pose a significant risk to human society, resulting in potential harm to property, infrastructure, injuries, and even loss of life~\cite{stephenson2008definition}.
Such events (e.g., heatwaves, floods, droughts) have experienced a noticeable surge in frequency, intensity, and duration in recent years due to climate change~\cite{machine2023ccarinoa}.
Significant concerns have been raised among human society. Appropriate measures and optimal strategies are urgently needed to forecast these events and mitigate their negative impacts. 
Considerable researchers have studied weather forecasting for years.
For example, \citet{kang2020prediction} used factors such as temperature, wind, and pressure data to forecast precipitation. Similarly, \citet{huang2021solar} relied on historical radiation data to forecast future radiation.

\hb{However, those works only focus on meteorological variables forecasting, missing forecasting textual weather and climate events or mapping the relationship between meteorology and its textual consequence.
In the real world, there exists a wide array of weather and climate events, also including their cascading secondary and tertiary consequences. For example, heavy rainfall can lead to waterlogging, infrastructure degradation (e.g., road collapse), traffic disruptions, and human casualties when urban systems lack adaptive resilience to climate change. A critical challenge lies in the disparity between heterogeneous numerical meteorological data and textual descriptions of such events and their consequences. The gap hinders the accurate forecasting of real-world textual events and their downstream impacts based solely on meteorological variables.}

\hb{To address this critical gap, we propose the \textbf{W}eather and \textbf{C}limate \textbf{E}vent \textbf{F}orecasting (WCEF), a novel task in environmental forecasting. Unlike traditional approaches that predict numerical meteorological variables (e.g., precipitation)~\cite{kang2020prediction, yang2024seasonality}, WCEF pioneers the direct generation of natural language descriptions to forecast weather and climate events and their cascading real-world consequences. This represents a fundamental paradigm shift, as it replaces the conventional workflow—where domain experts manually interpret numerical predictions to infer potential events, a process that is labor-intensive, subjective, and lacking geographical adaptability—with an automated, context-aware framework.}

\hb{The complexity of WCEF stems from its dual departure from existing methods: (1) its output requires generating textual event descriptions rather than numerical sequences, and (2) its input demands multimodal integration of numeric, text, and image. Prior efforts, which focus solely on single-modal numerical predictions, fail to address these challenges~\cite{schultz2021can}. They neither automate the translation of raw climate data into actionable narratives nor adapt to location-specific vulnerabilities, leaving critical gaps in forecasting precision and practical utility.}

\hb{
The emergence of LLMs and multimodal LLMs (MLLMs) presents significant potential for benchmarking and addressing the WCEF task.
To benchmark this task, we collect weather and climate events from environmental news articles, which serve as a valuable complement to meteorological data. These articles provide detailed descriptions of weather and climate events~\cite{michelle2023what}, including temporal and spatial information that can support precise event prediction. However, the sheer volume of unstructured environmental news conceals task-critical information. Additionally, the scarcity of supervised datasets, exacerbated by the novelty of the task and the complexity of curating aligned multimodal pairs (numerical data and event descriptions), requires significant domain expertise. The recent advancements in LLMs~\cite{brown2020language} offer a promising solution. They enable the extraction of structured information from vast amounts of unstructured documents~\cite{dagdelen2024structured, biswas2024robustness}. Specifically, we utilize GPT-4o-mini~\cite{gpt4o} to process 26,156 news articles, generating structured event representations (e.g., event A, cause, event B, location, and date) from unstructured text. To ensure accuracy, domain experts manually validate the outputs, resulting in the construction of a high-fidelity, spatiotemporally aligned multimodal dataset, CLLMate.

To address the WCEF task, bridging the modality gap between meteorological raster data (high-dimensional spatiotemporal grids) and textual descriptions remains a significant challenge~\cite{liu2023llava, jiang2024mantis}. 
MLLMs, in particular, offer an opportunity to integrate and align textual narratives with spatiotemporal data for the WCEF task~\cite{xu2023multimodal}.
To evaluate current MLLM's capabilities, we conduct extensive experiments on our constructed dataset using 23 MLLMs. The results demonstrate that while these models show potential, they remain constrained in their performance on the WCEF task. The experiment results reveal that current models have the advantage but are limited in the WCEF task. It underscores the need for further work to achieve accurate weather and climate event forecasting.
}

In summary, the contributions can be outlined:
\begin{itemize}[noitemsep,topsep=0pt,leftmargin=*]
    \item We propose the WCEF task to forecast textual events based on numerical meteorological raster data, which differs from traditional meteorological data forecasting. 
    \item \hb{We leverage domain knowledge extracted from the news corpus to establish the first multimodal dataset\footnote{The dataset will be publicly released.} to connect events and meteorology.}
    \item \hb{We conduct extensive experiments to evaluate existing MLLMs and our fine-tuned models on the WCEF task, benchmarking their capability in forecasting textual weather and climate events.}
\end{itemize}

%% file: 2-rw.tex
\section{Related Work}
Weather and climate forecasting has long been a research problem in the field of environment. In this paper, we proposed a more challenging task to predict textual events using MLLMs. 

\begin{table*}[ht!]
\small
\color{hbblue}
\centering
\setlength{\aboverulesep}{1pt}
\setlength{\belowrulesep}{1pt}
\makeatletter
\def\hlinewd#1{%
  \noalign{\ifnum0=`}\fi\hrule \@height #1 \futurelet
   \reserved@a\@xhline}
\makeatother
\newcolumntype{C}[1]{>{\centering\arraybackslash}p{#1}}
\begin{tabular}{cccc}
\hlinewd{0.8pt}
\toprule
\multirow{2}{*}{\textbf{Dataset}} & \multirow{2}{*}{\textbf{Meteorological Variables}} & \multicolumn{2}{c}{\textbf{Textual Events}} \\
\cmidrule(lr){3-4}
 & & \textbf{Not Post Records} & \textbf{Post Records} \\
\midrule
SEVIR~\cite{veillette2020sevir} & Satellite & $\times$ & $\times$ \\
Mesogeos~\cite{kondylatos2024mesogeos} & ERA5 & $\times$ & $\times$ \\
Digital Typhoon~\cite{kitamoto2024digital} & Satellite & $\times$ & $\times$ \\
Vaid et al.’s work~\cite{towards2022vaid} & $\times$ & $\times$ & Social Media \\
NeuralNERE~\cite{mishra2021neuralnere} & $\times$ & $\times$ & News \\
ClimateIQA~\cite{chen2024vision} & ERA5 & QA generated by LLM & $\times$ \\
WeatherQA~\cite{ma2024weatherqa} & ERA5 & Expert's pre-analysis & $\times$ \\
\textbf{CLLMate} & ERA5 & $\times$ & \textbf{News, verified by experts} \\
\bottomrule
\end{tabular}
\caption{\color{hbblue}Comparison with CLLMate with the existing datasets. CLLMate uniquely integrates ERA5 reanalysis data with expert-verified textual events, addressing the scarcity of aligned multimodal datasets for weather and climate event forecasting. (ERA5: the fifth generation ECMWF atmospheric reanalysis dataset of the global climate.)}
\label{tab:dataset}
\color{black}
\end{table*}

\subsection{Weather and Climate Events Forecasting}
In the era preceding modern weather prediction, human experience connects diverse natural signs, such as cloud patterns and animal behavior, with weather and their subsequent effects~\cite{risiro2012weather}.
The start of modern weather forecasting was marked with the first modern weather chart~\cite{allaby2009atmosphere, young2022history}. In contemporary study, a shift towards efficiency has occurred by consolidating various numerical variables into a unified numerical framework.
Two primary numerical methodologies are commonly employed: numerical weather prediction, which utilizes numerical simulation methods~\cite{bauer2015quiet, peter2008origins}, and AI-based forecasting, which leverages data-driven approaches~\cite{bi2023accurate, hewage2021deep}.
Both methods focus on numeric data, which experts are required to analyze to predict potential consequences. In contrast to those works, we make the first attempt to study the WCEF task of predicting textual events with deep learning models to ease expert labor. 

\subsection{Multimodal LLM} 
With the advancement of LLMs, there's been a surge of research on building multimodal LLMs. Their studies~\cite{wu2023next, zhan2024anygpt} try to incorporate multiple types of data beyond just natural language, such as images, audio, and video. BLIP-2~\cite{li2023blip} has developed a large-scale image captioning dataset, combining a language model with a vision encoder to create a multimodal model. Building on this, LLaVA~\cite{liu2023llava} introduces a more cost-effective method for training multimodal models through visual instruction tuning. 
The following multimodal LLMs, including QwenVL2~\cite{Qwen2-VL}, CogVLM~\cite{wang2023cogvlm}, deepseek-vl~\cite{lu2024deepseek}, Intern-vl~\cite{chen2024internvl, dong2024internlm}, etc., all follow a similar architecture of LLaVA. 
Following them, our work focuses on meteorology.

\hb{
\subsection{Existing Similar Dataset}
Prior efforts to model meteorological events fall into three categories (\autoref{tab:dataset}):

\textbf{Single-Modality Physical Datasets.}
Works like SEVIR (storm imagery)~\cite{veillette2020sevir}, Mesogeos (wildfire-related variables)~\cite{kondylatos2024mesogeos}, and Digital Typhoon (typhoon trajectories)~\cite{kitamoto2024digital} focus on meteorological measurements but lack textual event narratives. While valuable for physical modeling, they omit explicit linkages to real-world impacts.

\textbf{Text-centric datasets.}
Text-centric datasets such as Vaid et al.'s work (social media events)~\cite{towards2022vaid} and NeuralNERE (news-derived knowledge graphs)~\cite{mishra2021neuralnere} catalog climate-related events but fail to link them to corresponding meteorological data. This disconnect leaves the causal chain between weather patterns (e.g., extreme rainfall) and societal consequences (e.g., floods) unquantified.

\textbf{Multimodal Approaches with Limitations.}
Several other meteorological multimodal dataset construction efforts were conducted around the same time as CLLMate. However, they remain limited by data quality and scope.
ClimateIQA ~\cite{chen2024vision} pairs ERA5 data with LLM-generated QA pairs but lacks verified event records and impact descriptions. WeatherQA~\cite{ma2024weatherqa} aligns ERA5 with expert pre-analysis texts, yet it focuses on meteorological forecasts rather than post-event records and excludes societal impacts. Both datasets prioritize meteorological phenomena over downstream consequences and rely on synthetic or non-verified textual data, limiting their utility for impact-driven research.

CLLMate bridges these gaps by integrating expert-verified post-event records from authoritative news sources with spatiotemporally matched meteorological raster data. 
Unlike prior works, CLLMate (1) captures both meteorological phenomena (e.g., rainfall) and their downstream consequences (e.g., flooding, infrastructure damage), (2) establishes explicit cross-modal mappings between textual event descriptions (including time and location) and meteorological variables, and (3) ensures reliability through domain expert validation of event reports. 
}

%% file: 3-pre.tex
\section{Problem Formulation}
In this section, we provide an introduction to the data type in our dataset, CLLMate, and outline the formulation of the WCEF task (\autoref{fig:arch}).

\subsection{Spatio-Temporal Data}


Two types of spatio-temporal data are utilized: meteorological raster data and event data.
\subsubsection{Meteorological Raster Data}
Meteorological raster data is commonly employed in the environmental domain to depict meteorological variables. For instance, one global meteorological variable can be represented as a three-dimensional tensor $\boldsymbol{R} \in \mathbb{R}^{T\times\Lambda\times\Phi}$, where $T$ denotes the time dimension, and $\Lambda$ and $\Phi$ represent the longitude and latitude dimensions, respectively. Each element $\boldsymbol{r}_{t,\lambda, \phi}$ corresponds to the variable value at a specific time $t$ and location $(\lambda, \phi)$.
\subsubsection{Event Data}
On the other hand, event data $\boldsymbol{E}$ are occurrences of interest at specific points in space and time with the shape of $T\times\Lambda\times\Phi$. For example, a single event can be represented using textual information $\boldsymbol{e}_{t,\lambda, \phi}$, indicating the occurrence at a particular time $t$ and location $(\lambda, \phi)$. 

Two distinctions exist between raster and event data. First, raster data is numerical, while event data is textual. Next, raster data is dense, whereas event data is sparse, featuring specific instances at discrete locations and times.

\subsection{WCEF Task}
\begin{figure}[t]
  \centering
  \includegraphics[width=\linewidth]{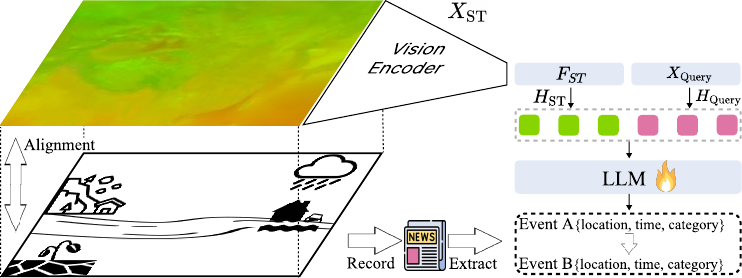}
  \caption{The CLLMate framework is designed to extract textual weather and climate events and align them with meteorological raster data for the WCEF task.}
  \label{fig:arch}
\end{figure}

The existing weather forecasting task involves predicting the future target meteorology variable using either the same or more variables. This task can be formulated as follows:
\begin{equation}
    \setlength{\abovedisplayskip}{3pt}
    \setlength{\belowdisplayskip}{3pt}
    \boldsymbol{R}_{C+1:C+F}=f(\boldsymbol{R}_{C-H+1:C})
\end{equation}

\hb{where $f(\cdot)$ is the forecasting model, $\boldsymbol{R}_C$ denotes the numerical meteorological variables at the current time, F represents the length of the forecasting data, and H is the length of historical data used.}

In contrast, the objective of the proposed WCEF task is to predict textual events using numerical data (\autoref{fig:arch}). 
These events can be defined based on meteorological characteristics, such as heatwaves and the subsequent consequences they entail. 
The scope of events encompasses primary occurrences like heatwaves and droughts, as well as secondary and tertiary events (consequences) like landslides and human casualties that arise as a result of the primary events.
These examples illustrate the complexity of deriving such events solely from numerical data. It highlights the significance of the historical events and the valuable knowledge they offer.
The WECF task can be formulated:
\begin{equation}
    \setlength{\abovedisplayskip}{3pt}
    \setlength{\belowdisplayskip}{3pt}
    \boldsymbol{E}_{C:C+F-1}=f_{\phi}(\boldsymbol{R}_{C-H+1:C}, \hspace{2pt} \text{Instruction})
\end{equation}
\hb{where $f_{\phi}(\cdot)$ denotes the LLM model and $\boldsymbol{E}_C$ is the textual events happening at the current time. It is worth emphasizing that, unlike existing forecasting tasks utilizing the same source data, we employ C in both the forecasting and historical data. This is because $\boldsymbol{E}_C$ and $\boldsymbol{R}_C$ hold distinct meanings.}


%% file: 4-dataset.tex
\section{Dataset Construction Process}

\begin{figure}[t]
  \centering
  \includegraphics[width=\linewidth]{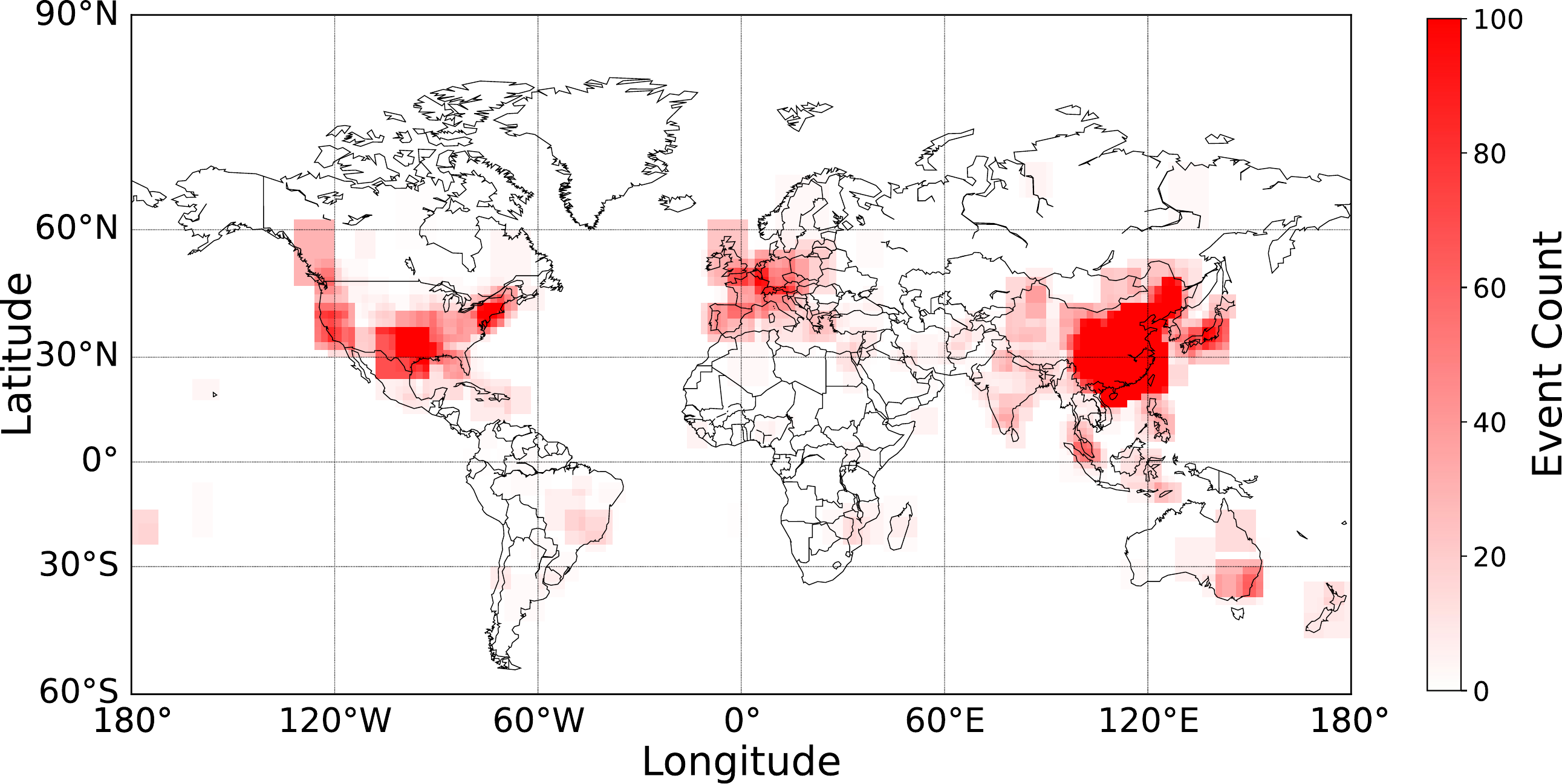}
  \caption{\color{hbblue}Spatial distribution of extracted events. Each rectangle represents an extracted event. The events span most global regions, with notable concentrations in East Asia, North America, and Europe.}
  \label{fig:spatial}
\end{figure}
\begin{figure}[t]
  \centering
  \includegraphics[width=\linewidth]{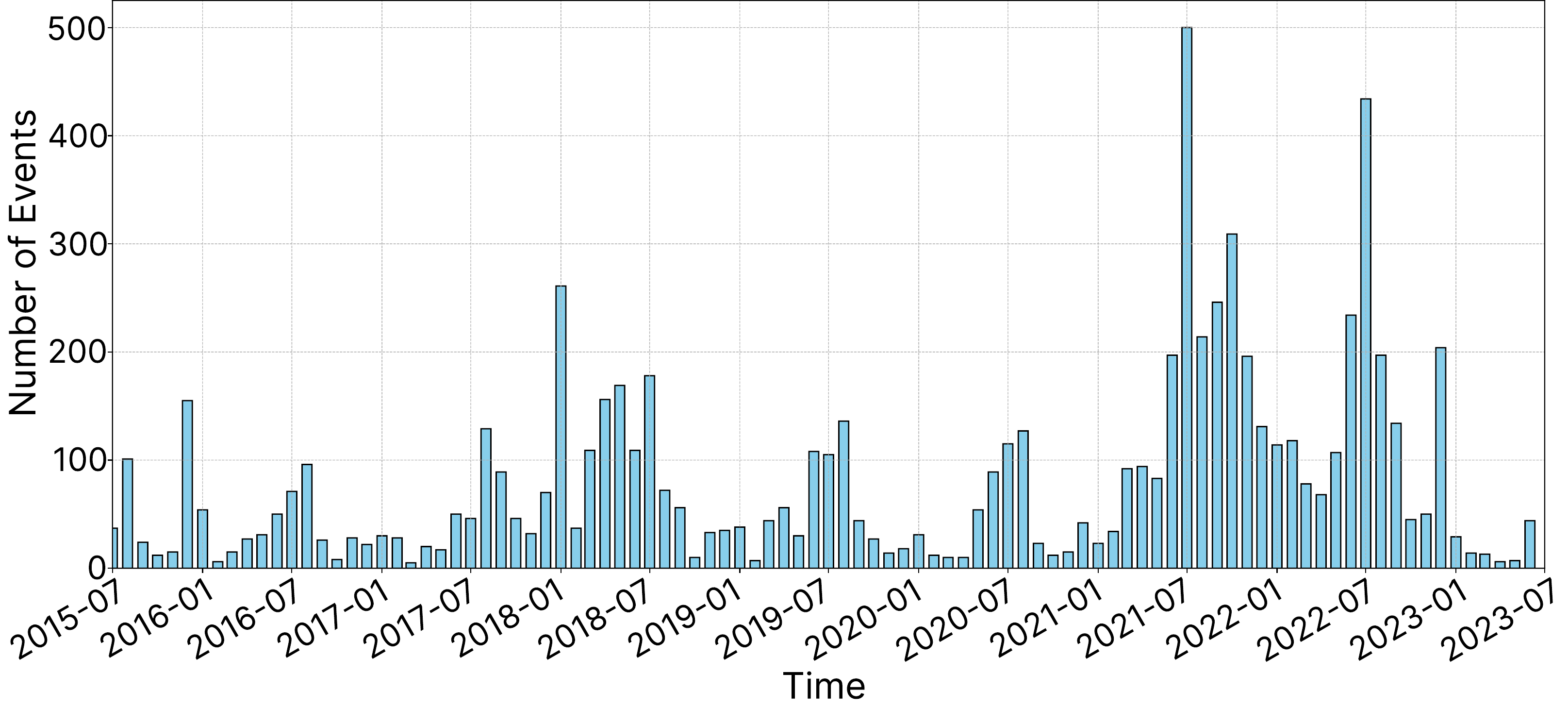}
  \caption{\color{hbblue}Temporal distribution of extracted events. The events span a long time period, from 2015 to 2023. A notable outlier in the number of events occurred due to the catastrophic flooding in Zhengzhou in July 2021.}
  \label{fig:temporal}
\end{figure}
In this section, we introduce the meteorological raster dataset and environmental news dataset we used to construct the multimodal instruction dataset for the WCEF task. Then, we explain the pipeline to create the multimodal dataset (\cref{data_format}).

\subsection{Dataset Used to Construct CLLMate}

\para{Meteorological Raster Dataset.}
We use the open and free ERA5 reanalysis dataset~\cite{era52020hersbach} to obtain the meteorology data (Copernicus License). Its usage in climate research has been widely acknowledged for its quality~\cite{increased2022sun}.
Four variables, namely ``2m temperature,'' ``10m u-component of wind,'' ``10m v-component of wind,'' and ``total precipitation,'' were selected to forecast the following textual weather and climate events. 
The dataset consists of hourly data spanning from July 2015 to July 2023, with a spatial resolution of $27.75\ km \times 27.75\ km$. 

\para{Environmental News Dataset.} 
We acquired environmental news from Wisers~\cite{wisers2024} through a procurement process, which consisted of highly environment-related news articles from news publishers. The dataset spans from July 2015 to June 2023. 
Each news article contains the title, content, character statistics, date, publisher, and media type.
The media types encompass both web and publication resources while excluding internet-based media sources primarily reliant on aggregating news reports from official news agencies.


\hb{
\subsection{The Multimodal Dataset}
We construct the first multimodal dataset for the WCEF task through three steps.

\subsubsection{Event Extraction} 
\label{sec:extract}
Weather and climate events exist within a vast amount of news articles. 
Research has demonstrated the efficacy of utilizing LLMs for extracting structured information from text~\cite{dagdelen2024structured}.
The initial step of our pipeline is to identify and extract these events and their relationships from each news document. 
First, we employed GPT-4o-mini to analyze 26,156 environmental news articles, extracting weather and climate event knowledge in structured triples with spatiotemporal metadata (e.g., high temperature causes heatwave, Hong Kong, 2022-08-16). The details of the prompt for extracting events can be found in \cref{extract_prompt}.

Initial automated labeling identified 6,352 news containing potential event relationships. These articles, along with their extracted triples, locations, and dates, underwent rigorous manual validation by three meteorological domain experts. 
Events, which are not meteorological events, or lack clear spatial information, or temporal details will be excluded.
Following verification, 2,575 news were confirmed to contain accurate event knowledge, yielding a final curated dataset of 7,747 spatiotemporally grounded events. 
Please refer to \cref{annotation} for more details on annotation.

Specially as for the spatial aspect, there are 1,049 distinct regions in the dataset. To derive geospatial bounds for these locations, we continue to employ GPT-4o-mini to generate latitude-longitude coordinates, which are subsequently validated by humans to ensure alignment with real-world geographical boundaries by visualizing them on the map. Additionally, we expand the geographical range, with a minimum of 5 degrees, to incorporate surrounding meteorological conditions.

\para{Spatial Distribution.}
The dataset exhibits a broad geographical representation, covering most global regions (\autoref{fig:spatial}). Events are notably concentrated in East Asia, North America, and Europe, with comparatively fewer instances in other regions. This imbalance likely reflects socioeconomic disparities in environmental reporting infrastructure and media coverage biases. 

\para{Temporal Distribution.}
The temporal coverage spans July 2015 to July 2023, with marked seasonal patterns: event frequency peaks during summer and winter months (\autoref{fig:temporal}). A significant outlier occurred in July 2021, coinciding with catastrophic flooding since extreme rainfall in Zhengzhou, China, which caused hundreds of fatalities and widespread infrastructure damage. This event garnered exceptional media attention.

\begin{figure}[t]
  \centering
  \includegraphics[width=\linewidth]{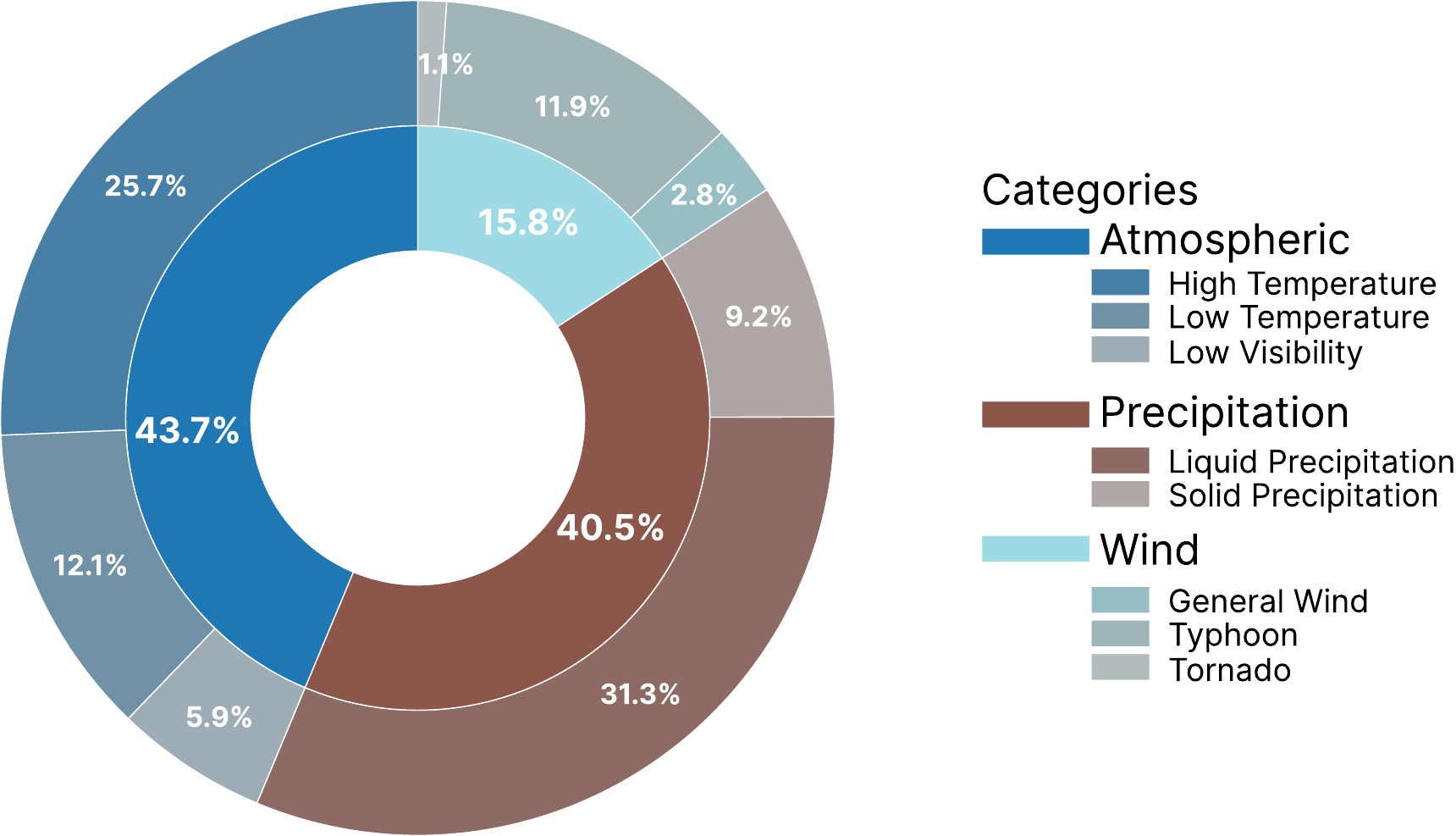}
  \caption{\color{hbblue}Distribution of categories within the meteorological phenomena category (3,979/7747 events). The distribution is imbalanced, reflecting the nature of event reporting in the news.}
  \label{fig:groupA}
\end{figure}
\begin{figure}[t]
  \centering
  \includegraphics[width=\linewidth]{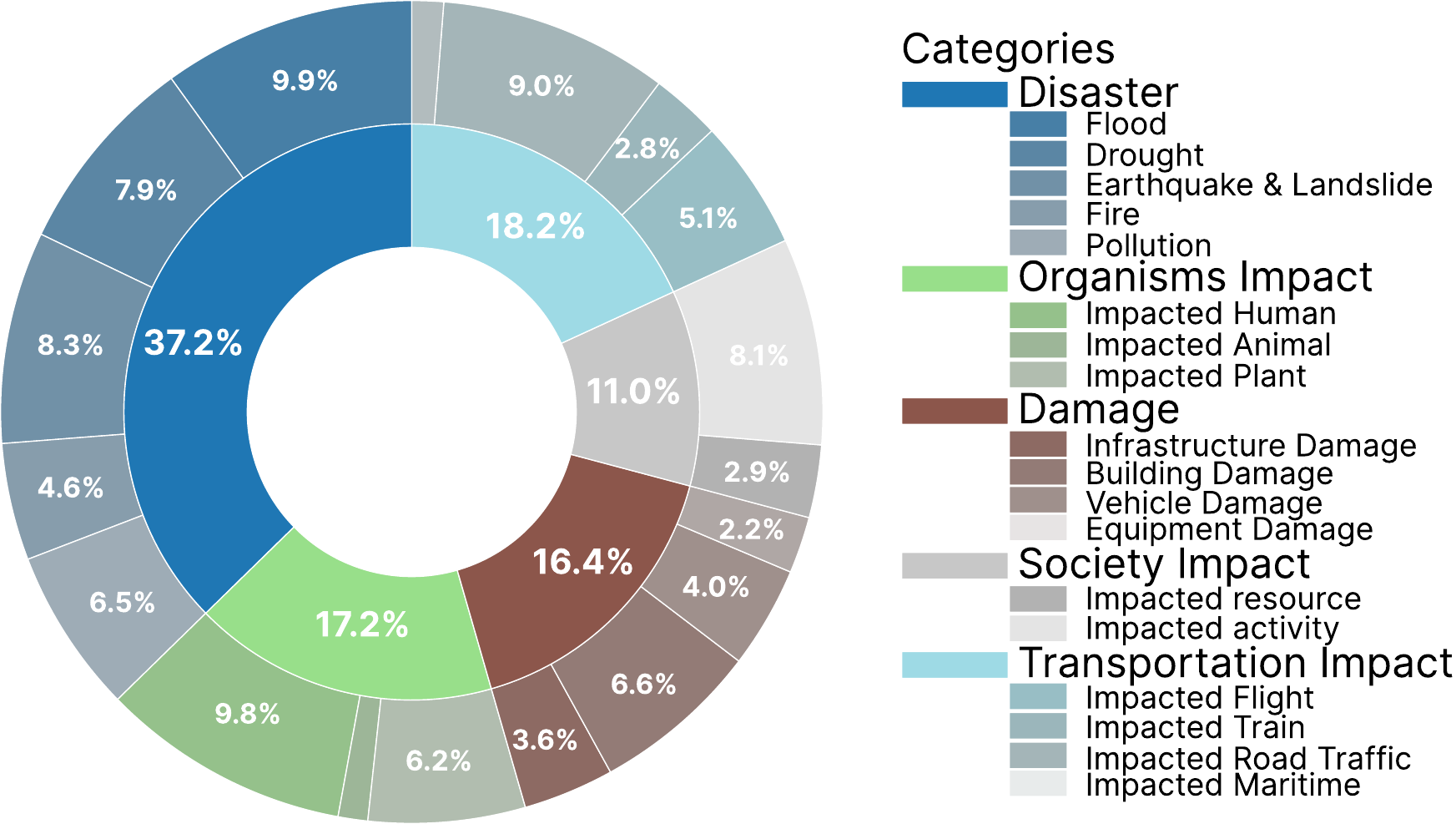}
  \caption{\color{hbblue}Distribution of categories within the consequences category (3,768/7747 events).  The distribution is imbalanced, reflecting the nature of event reporting in the news.}
  \label{fig:groupB}
  \vspace{-6pt}
\end{figure}

\subsubsection{Event Categorization}
To simplify the evaluation of MLLMs’ capabilities in the WCEF task, collaborating with domain experts, we design the problem into a multiple-choice framework by categorizing weather and climate events hierarchically (\cref{category}). 
The categorization combines a bottom-up approach to clustering (using semantic similarity) with a top-down approach (using the taxonomy in the environmental domain)~\cite{gaiteri2015identifying}.
Events are classified into two primary categories: meteorological phenomena (3979/7747 events, \autoref{fig:groupA}) and their cascading consequences (3768/7747 events, \autoref{fig:groupB}). Each category is further divided into fine-grained subclasses, which serve as candidate options for model selection.
In the benchmarking, MLLMs are required to select the most probable predefined fine-grained subclass.
We split the dataset into train and test sets. For the category distribution of train and test sets, please refer to \cref{distrubution}.


The statistical distributions of these subclasses are visualized in Figures~\autoref{fig:groupA} and~\autoref{fig:groupB}. There are imbalances in the category distribution, such as for the primary category of meteorological phenomena, there are 43.7\% events are atmospheric-related, 40.5\% events are precipitation-related, and only 15.8\% events are wind-related. That is because all events are extracted from the news and follow the distribution of the recording of events.
}

\hb{
\subsubsection{Event-Meteorology Alignment}
Multimodal dataset construction relies on precise spatiotemporal alignment between events and their meteorological conditions (\autoref{fig:arch}). 
For each event (\cref{data_format}), we integrate: (1) textual event, (2) meteorological raster data for the event’s specific date and region, and (3) regional statistical context (e.g., max) from the same spatiotemporal origin.
Meteorological inputs are preprocessed into RGB-like grids (normalized to [0, 1]):
\begin{itemize}[noitemsep,topsep=0pt,leftmargin=*]
    \item Channel 1: 2m temperature.
    \item Channel 2: 10m wind speed (u- and v-components combined via magnitude norm).
    \item Channel 3: total precipitation.
\end{itemize}
This representation enables MLLMs to process spatial climate patterns as visual inputs.
}


%% file: 5-benchmark.tex
\section{Benchmarking Results}
\hb{\subsection{Benchmarked Models}
The dataset is partitioned chronologically into training (90\%) and testing (10\%) sets. We conducted extensive experiments with 23 MLLMs on the test set, encompassing closed-source, open-source, and our fine-tuned models:}

\begin{table*}[ht!]
\color{hbblue} 
\centering
\setlength{\aboverulesep}{1pt}
\setlength{\belowrulesep}{1pt}
\small
\makeatletter
\def\hlinewd#1{%
  \noalign{\ifnum0=`}\fi\hrule \@height #1 \futurelet
   \reserved@a\@xhline}
\makeatother
\newcolumntype{C}[1]{>{\centering\arraybackslash}p{#1}}
\begin{tabular}{ p{6.2cm} | C{1.8cm}C{1.8cm} | C{2cm}C{2cm}}
\hlinewd{0.8pt}
\toprule
\textcolor{hbblue}{\vfill\textbf{Models}} & \textcolor{hbblue}{\small{Primary Category of Phenomena}} & \textcolor{hbblue}{\small{Subordinate Category of Phenomena}} & \textcolor{hbblue}{\small{Primary Category of Consequence}} & \textcolor{hbblue}{\small{Subordinate Category of Consequence}} \\[1mm] 
\midrule
Random guess  & 33.33\% & 12.50\% & 20.00\% & 5.56\% \\ 
Majority guess  & 40.12\% & 26.07\% & 40.11\% & 4.90\% \\ 
\midrule
GPT-4o-mini~\cite{gpt4o} & 30.28\% & 13.62\% & 34.63\% & 14.13\% \\
GPT-4o~\cite{gpt4o} & 32.11\% & 15.65\% & 39.22\% & 12.01\% \\ 
GPT-o1~\cite{gpt4o} & 32.93\% & 10.97\% & 17.67\% & 10.60\% \\ 
Gemini-1.5-pro~\cite{team2023gemini} & 20.12\% & 8.74\% & 28.98\% & 10.25\% \\ 
Gemini-2.0-flash~\cite{team2023gemini} & 37.80\% & 19.72\% & 30.39\% & 13.43\% \\ 
Claude-3.5-Sonnet~\cite{claudesonnet} & 37.60\% & 19.72\% & 40.28\% & \underline{\textbf{15.90\%}} \\ 
\midrule
Janus-Pro-7B~\cite{chen2025janus} & 31.31\% & 2.64\% & 21.20\% & 2.83\%  \\ 
InternVL2-8B~\cite{chen2024internvl} & 29.27\% & 10.37\% & 21.91\% & 5.30\%   \\
Cambrian-1-8B~\cite{tong2024cambrian} & 31.10\% & 9.56\% & 31.80\% & 4.59\%   \\
LLaVA-1.5-7B~\cite{liu2023llava} & 31.30\% & 14.02\% & 20.85\% & 3.89\%  \\ 
LLaVA-1.6-vicuna-7B~\cite{liu2024llavanext} & 31.30\% & 13.82\% & 20.85\% & 3.89\%   \\
LLaVA-1.6-mistral-7B~\cite{liu2024llavanext} & 32.93\% & 17.28\% & 39.22\% & 12.01\%  \\ 
LLaVA-1.5-13B~\cite{liu2023llava} & 31.32\% & 15.04\% & 28.27\% & 7.77\%  \\ 
LLaVA-1.6-vicuna-13B~\cite{liu2024llavanext} & 29.47\% & 14.63\% & 21.91\% & 7.42\%   \\
LLaVA-1.6-vicuna-34B~\cite{liu2024llavanext} & 42.28\% & 15.45\% & 17.67\% & 3.89\%   \\
Qwen2-VL-7B~\cite{wang2024qwen2} & 32.72\% & 11.38\% & 25.80\% & 5.30\%  \\ 
Qwen2.5-VL-7B~\cite{Qwen2.5-VL} & 42.27\% & 17.48\% & 21.91\% & 7.78\%  \\ 
Qwen2.5-VL-72B~\cite{Qwen2.5-VL} & 40.85\% & 20.73\% & 39.58\% & \textbf{16.25\%}  \\ 
\midrule
Fine-tuned-LLaVA-1.5-7B & \textbf{45.93\%} & 23.37\% & 43.46\% & 9.89\%  \\ 
Fine-tuned-LLaVA-1.6-vicuna-7B & 44.91\% & \textbf{29.67}\% & \textbf{44.17\%} & 8.48\%  \\ 
Fine-tuned-LLaVA-1.6-mistral-7B & 45.53\% & 25.61\% & \underline{\textbf{43.82\%}} & 8.13\%  \\ 
Fine-tuned-LLaVA-1.5-13B & 44.92\% & 23.78\% & 43.46\% & 6.36\%  \\ 
Fine-tuned-LLaVA-1.6-vicuna-13B & \underline{\textbf{45.54\%}} & \underline{\textbf{27.03\%}} & 42.40\% & 9.19\%  \\ 
\bottomrule
\end{tabular}
\caption{\color{hbblue}The accuracy of benchmarked MLLMs across two granularity levels: primary category classification and subordinate category identification for both meteorological phenomena and their cascading consequence.}
\label{tab:performance}
\color{black} 
\end{table*}

\hb{
\subsubsection{Closed-Source Models}
\begin{itemize}[noitemsep,topsep=0pt,leftmargin=*]
    \item \textbf{GPT Series}~\cite{gpt4o}: OpenAI’s flagship MLLM with integrated vision capabilities. We evaluated three variants of OpenAI's GPT architecture optimized for multimodal reasoning: GPT-4o, GPT-4o-mini, and GPT-o1.

    \item \textbf{Gemini}~\cite{team2023gemini}: Google's MLLM with enhanced context window and optimized for high-speed processing of text-image inputs. We evaluated Gemini-1.5-pro and Gemini-2.0-flash.

    \item \textbf{Claude-3.5-Sonnet}~\cite{claudesonnet}: Anthropic's MLLM with improved analytical capabilities, supporting text and image inputs. 
\end{itemize}

\subsubsection{Open-Source Models}
\begin{itemize}[noitemsep,topsep=0pt,leftmargin=*]
    \item \textbf{Deepseek-Janus-Pro-7B}~\cite{chen2025janus}: Deepseek's MLLM with both multimodal understanding and text-to-image capabilities.

    \item \textbf{InternVL2-8B}~\cite{chen2024internvl}: A vision-language foundation model with powerful capabilities in handling complex multimodal data.

    \item \textbf{Cambrian-1-8B}~\cite{tong2024cambrian}: A fully open multimodal LLM designed with a vision-centric approach.

    \item \textbf{LLaVA-1.5/1.6}~\cite{liu2023llava, liu2024llavanext}: Versions of the widely adopted open-source MLLM framework: LLaVA-1.5-7B/13B: MLLMs with standard vision-language alignment. LLaVA-1.6-7B/13B/34B: Enhanced variants MLLMs for fine-grained visual grounding.

    \item \textbf{Qwen2/2.5}~\cite{wang2024qwen2} utilize dynamic resolution and frame rate training for video understanding. We evaluate Qwen2-VL-7B and Qwen2.5-VL-7/72B.
\end{itemize}

\subsubsection{Fine-tuned Models}
We fine-tuned a series of LLaVA models using the training set of the CLLMate to evaluate the value of the dataset and the efficiency of MLLMs in solving the WCEF task. The fine-tuned models include \textit{LLaVA-1.5-7B/13B} and \textit{LLaVA-1.6-vicuna-7B/13B}. They are trained on 8 A800 GPUs for one epoch. Please refer to \cref{training} for more details.

\subsection{Prompt Design}
We design specialized prompts for MLLMs to forecast meteorological phenomena and their consequences. 
The prompts consist of ``System Prompt,'' ``Location Information,'' ``Meteorological Parameters,'' ``Meteorological Image,'' ``Options,'' and ``Zero-Shot CoT~\cite{kojima2022large}.'' 
As for the Zero-Shot CoT, we use (1) analysis of statistical data, (2) analysis of spatial patterns in the image, and (3) synthesis of findings to select the most probable fine-grained category.
To ensure precision and minimize ambiguity, outputs are constrained to standardized serial codes (e.g., A.1 high temperature.)
Additionally, to mitigate positional bias, answers are randomized across instances (e.g., "high temperature" may appear as Option A.1 in one instance and Option B.2 in another). 
Refer to \cref{phenomena_prompt} and \cref{consequence_prompt} for more details.}


\hb{
\subsection{Evaluation Metric}
For the benchmarking of multi-choice questions, we employ accuracy as the metric, following the previous research, such as Ge et al.'s work~\cite{ge2022bridging}. Accuracy is a straightforward metric that measures the proportion of correct answers out of the total number of questions. 

\subsection{Heuristic Baselines}
We implement two heuristic baselines: 
\begin{itemize}[noitemsep,topsep=0pt,leftmargin=*]
    \item Random Guess: Answers are selected uniformly at random from all valid options, reflecting chance-level performance.
    \item Majority Guess: The most frequent category/subcategory in the training set (Precipitation/Liquid Precipitation and Disaster/Flood, \autoref{fig:train}) is chosen for all test instances. Thus, the distribution of them in the test set will be used (40.12\%/26.07\% and 40.11\%/4.90\%).
    It serves two purposes: (1) quantifying the advantage of leveraging dataset biases, and (2) contextualizing model performance relative to empirically informed predictions.
\end{itemize}

}

\hb{
\subsection{Main Results}
Table~\ref{tab:performance} summarizes the performance of benchmarked MLLMs across two granularity levels: primary category (coarse-grained) and subordinate category (fine-grained) for meteorological phenomena and their consequences. The results are also shown in \autoref{fig:A} and \autoref{fig:B}.

\subsubsection{Meteorological Phenomena Forecasting}
\textbf{Primary Category}: Most closed-source and open-source models perform near random guess (30.28–37.80\%), with the exception of LLaVA-1.6-vicuna-34B and Qwen2.5 (40.85\%-42.28\%), which exceeds the majority guess baseline (40.12\%). Fine-tuned models achieve substantial gains, with Fine-tuned-LLaVA-1.5-7B attaining the highest accuracy (45.93\%).

\textbf{Subordinate Category}: While most existing models surpass random guessing, none exceed the majority guess baseline (26.07\%). Fine-tuning yields significant improvements (e.g., Fine-tuned-LLaVA-1.6-vicuna-7B: 29.67\% vs. 13.82\% for its untrained counterpart).

\subsubsection{Consequence Forecasting}
\textbf{Primary Category}: Claude-3.5-Sonnet leads among untrained models (40.28\%), while fine-tuned variants dominate, notably Fine-tuned-LLaVA-1.6-vicuna-7B (44.17\%), surpassing the random guess (20.00\%) and majority guess baseline (40.11\%). This demonstrates the learnability of meteorology-to-impact mappings when models are task-adapted.

\textbf{Subordinate Category}: QWen2.5-VL-72B achieves the highest accuracy (16.25\%), marginally exceeding the heuristic baselines (5.56\%/4.90\%), while fine-tuned models show limited gains.
}

%% file: 6-analysis.tex
\hb{\section{Performance Analysis}
We show some findings of the benchmarking. For more analysis, such as case study, ablation study, and scale analysis, please refer to \cref{analysis}.
\subsection{MLLMs vs. Heuristic Baselines}
While MLLMs outperform the random guessing baseline, some of them often fall short of the majority guess baseline, a pattern attributable to an inherent class imbalance in the dataset. The imbalance provides a competitive advantage for naive reliance on prior knowledge of the category distribution. However, fine-tuned models such as Fine-tuned-LLaVA-1.5-7B surpass the majority guess baseline in phenomena prediction (45.93\% vs.40.12\%) and in consequence forecasting (43.46\% vs. 40.11\%). Critically, this demonstrates that MLLMs can learn meteorology-to-event mappings, achieving gains through pattern recognition.

\subsection{Closed-Source and Open-Source Models}
Among closed-source models, Claude-3.5-Sonnet achieves the highest forecasting accuracy. Surprisingly, GPT-o1's performance is lower than that of other models. 
Open-source models exhibit significant performance fragmentation. While LLaVA-1.6-34B achieves competitive primary phenomena accuracy (42.28\%), its consequence forecasting decreases to 17.67\%, worse than random guessing (20\%). Qwen2.5-VL-72B shows an advantage in consequence forecasting (39.58\%/16.25\%).
The performance of the best closed-source model is comparable to that of the best open-source model.

\subsection{Effectiveness of Task-Specific Fine-tuning}
Fine-tuning on CLLMate yields dramatic improvements:
Fine-tuned-LLaVA-1.5-7B surpasses Claude-3.5-Sonnet in primary phenomena accuracy (45.93\% vs. 37.60\%) and consequences (43.46\% vs. 40.28\%).
Subordinate category performance improves by 9–15 percentage points for phenomena (e.g., 23.37\% vs. 14.02\% in LLaVA-1.5-7B), though subordinate consequence forecasting remains challenging (<10\% accuracy).
This demonstrates that while MLLMs lack inherent weather reasoning capabilities, task-specific adaptation enables competitive forecasting. This underscores the necessity of task-specific alignment for bridging numerical meteorology and textual narratives.

\subsection{Limited Performance}
while existing or fine-tuned MLLMs outperform the heuristic baselines, they have relatively low accuracy (< 50\%), suggesting current models have the advantage, however, limited, in distinguishing different coarse-grained event types (e.g., Precipitation vs. Wind) and fine-grained event types (e.g., General wind vs. Typhoon). It underscores the need for future work to forecast meteorological events more accurately. 
}

%% file: 7-conclusion.tex
\hb{\section{Conclusion}
We propose the WCEF task, which forecasts weather and climate events using meteorological raster data. To support this task, we present CLLMate, the first supervised multimodal dataset that aligns spatiotemporally grounded event descriptions and their consequences with corresponding meteorological numerical data.
We conducted extensive experiments to assess 23 MLLMs and our fine-tuned MLLMs's performance on the dataset, CLLMate.
The results show that while the MLLMs' performance can outperform the heuristic baselines, they are still limited.
CLLMate establishes a foundational benchmark for WCEF, revealing both the feasibility of event forecasting and the pressing need for domain-optimized architectures.
}

%% file: 8-limitation.tex
\section*{Limitation}
We identify some limitations exist in this study and future research directions in this emerging field:


First, the inclusion of additional modalities to offer supplementary insights for forecasting is essential.
While this study initially delves into modalities encompassing numerical spatiotemporal data and text, there exist other meteorological data modalities like satellite imagery and time series data~\cite{jin2023time} from global meteorological stations that could contribute to longer-term, more precise forecasting. 
These data can be easily integrated into the events in CLLMate according to the spatiotemporal information.
Further research is warranted to enrich the forecasting capabilities.

Second, since we extracted a large number of triples (event A causes event B),
the incorporation of the knowledge graph necessitates additional exploration. 
Employing more sophisticated techniques to equip the MLLM with prior world knowledge~\cite{do2024constraintchecker, wang2024abspyramid} of weather and climate for the events understanding~\cite{wang2022subeventwriter, wang2023cola}, could potentially enhance the performance.